\documentclass[12pt]{article}

\usepackage[utf8]{inputenc} 
\usepackage{hyperref}       
\usepackage{url}            
\usepackage{booktabs}       
\usepackage{amsfonts}       
\usepackage{nicefrac}       
\usepackage{microtype}      
\usepackage{xcolor}         
\usepackage{algorithm}
\usepackage{algpseudocode}
\usepackage{subcaption}
\usepackage{amsmath}
\usepackage{comment}
\usepackage{natbib}
\usepackage{enumitem}
\usepackage{graphicx}
\usepackage{float}
\usepackage{placeins}
\usepackage{microtype}
\usepackage{multirow}
\usepackage{amsmath,amssymb,amsthm,mathtools,bm}
\usepackage{graphicx}
\usepackage{cleveref}
\usepackage{booktabs}     
\usepackage{caption}
\usepackage{tcolorbox}
\usepackage{appendix}
\usepackage{array}
\usepackage{xr}
\providecommand{\norm}[1]{\left\lVert #1 \right\rVert}

\newtheorem{theorem}{Theorem}
\newtheorem{corollary}{Corollary}
\newtheorem{definition}{Definition}

\newtheorem{condition}{Condition}

\newtheorem{proposition}{Proposition}

\graphicspath{{qss_figs/}}

\providecommand{\R}{\mathbb{R}}
\providecommand{\E}{\mathbb{E}}
\providecommand{\Var}{\operatorname{Var}}
\providecommand{\Prob}{\mathbb{P}}
\providecommand{\ind}{\mathbf{1}}
\providecommand{\eps}{\varepsilon}
\providecommand{\bstar}{\beta^\star}
\providecommand{\gtilde}{\widetilde g}

\makeatletter
\providecommand{\qssonlineappref}[2]{%
  \@ifundefined{r@S-#1}{Online Appendix~#2}{\csname ref\endcsname{S-#1}}%
}
\providecommand{\qssonlinelabeledref}[3]{%
  \@ifundefined{r@S-#1}{Online Appendix~#3}{#2~\csname ref\endcsname{S-#1}}%
}
\makeatother

\providecommand{\appComputational}{Appendix~\ref{app:simulation-supplement}}

\begin{document}

\def\spacingset#1{\renewcommand{\baselinestretch}%
{#1}\small\normalsize} \spacingset{1}

\date{}

  \title{\bf Locally Private Online Quantile Regression: Estimation and Inference}
  \author{Yi Liu\footnotemark[1] and Qirui Hu\footnotemark[2] \footnotemark[3] \hspace{.2cm}}
  \maketitle
  \renewcommand{\thefootnote}{\fnsymbol{footnote}}

  \footnotetext[1]{York University}
  \footnotetext[2]{Shanghai University of Finance and Economics}
  \footnotetext[3]{Corresponding author: huqirui@mail.shufe.edu.cn}

\bigskip

\begin{abstract}
We study estimation and inference for online quantile regression under a one-report user-level \(\eps\)-locally differentially private ($\eps$-LDP) protocol. The main difficulty is that the standard quantile-regression estimating-equation contribution couples covariates with a residual comparison, so a server that receives only privatized reports cannot form the usual online update. We address this by developing a finite-alphabet channel in which each user computes the contribution locally, applies support-aware stochastic quantization and randomized response to one selected-block category, and sends one report. A public decoder corrects the randomized-response distortion and reconstructs a server-side estimating-equation input with the correct conditional mean. These decoded inputs are then used in projected Polyak-Ruppert averaging. For fixed finite channel designs, we establish local privacy, decoder unbiasedness, consistency, asymptotic normality, and Hessian-free self-normalized inference for prespecified scalar contrasts. Simulations and a New York City taxi-trip illustration show that the private trajectory approaches the nonprivate online reference as the privacy budget grows and outperforms direct Laplace and face-exponential geometric releases in the reported regimes.
\end{abstract}

\vspace{0.5cm}

\newpage
\spacingset{1.2}

\section{Introduction}
\label{sec:introduction}

Quantile regression (QR) estimates conditional quantiles, or their linear-projection analogues under possible misspecification, by minimizing the asymmetric check loss of \citet{koenker1978regression}; see also \citet{koenker2005quantile,angrist2006quantile}. It is useful when conditional means give an incomplete distributional summary, for example under skewness, tail heterogeneity, or quantile-specific covariate effects. In modern data streams with sensitive records, the privacy risk is not limited to transmitting \(\left(X,Y\right)\): deterministic functions of a participant's record can also reveal sensitive attributes or participation. This motivates randomizing participant-side information before any server-side computation.

Local differential privacy (LDP) formalizes this trust model. Unlike central differential privacy, where a trusted curator first observes the database and then releases a privatized output \citep{dwork2006calibrating}, LDP requires each user to randomize the record before transmission \citep{warner1965randomized,duchi2013local,duchi2018minimax}. Thus the server observes only a randomized report whose distribution is stable, up to the factor \(e^\eps\), under changes to the participant's contribution. We study fixed-dimensional linear QR with bounded covariates under a one-record, one-report user-level \(\eps\)-LDP protocol; Definition~\ref{def:seq-ldp} gives the formal privacy condition. The server may broadcast any public or predictable query information, including the current online iterate and a public quantization design.

This setting creates an obstruction that is absent from ordinary online QR. Mathematically, at quantile level \(\tau\), a standard online update is driven by the estimating-equation contribution
\[
 g_\beta\left(X,Y\right)=X\left\{\ind\left(Y\le X^\top\beta\right)-\tau\right\}.
\]
The server needs both the covariate vector and the residual comparison \(\ind\left(Y\le X^\top\beta\right)\) to form this vector, but under user-level \(\eps\)-LDP it observes neither directly. Thus the problem is not simply to perturb an available stochastic gradient: the privatized report must protect the participant's contribution while its decoded version retains the conditional mean structure required by stochastic approximation.

Existing methods cover adjacent settings but not this one-report locally private estimating-equation problem. Nonprivate online QR assumes an observable record-level update \citep{shen2025onlineqr}; scalar locally private quantile methods do not involve covariate-dependent regression updates \citep{liu2023online,liu2024cdf,cai2025timeuniform,aamand2025lightweight,caiprivacy,hucensoring}; and central or distributed private QR relies on different trust or aggregation models \citep{chen2023dpquantileloss,tran2024dpqr,shen2025highdim,lu2025versatile}. We therefore formulate locally private online QR as a decoded estimating-equation problem in which LDP reports satisfy user-level \(\eps\)-LDP and decoded server inputs preserve the mean needed for averaged stochastic approximation. Three questions guide the development.
\begin{tcolorbox}[colback=white]
First, can a single data-dependent LDP report be decoded into an unbiased QR estimating-equation input at every public iterate? \\
Second, how does the finite randomized-response alphabet affect the effective information available to the online recursion? \\
Third, can confidence statements be built from the decoded private trajectory without estimating the QR Hessian?
\end{tcolorbox}


To answer these questions, we construct a support-aware finite-alphabet channel for decoded estimating equations. At each iterate, the user computes \(g_\beta\left(X,Y\right)\) locally. The QR residual comparison places this vector on one of two support faces, and bounded covariates restrict selected coordinates to public intervals. The user stochastically quantizes selected coordinates on a public grid, applies randomized response to the resulting finite category, and sends only the LDP report. The server applies an affine randomized-response decoder and Horvitz-Thompson coordinate reconstruction. Writing \(\widetilde g_\beta\) for the decoded server-side input, the key property is conditional unbiasedness:
\[
 \E\left\{\widetilde g_\beta\left(X,Y\right)\mid X,Y,\beta\right\}=g_\beta\left(X,Y\right).
\]
Privacy is a property of the LDP report received by the server; estimation and inference are post-processing operations applied to the decoded trajectory.

The paper makes the following contributions. \begin{enumerate}[leftmargin=*] \item We formulate streaming linear QR under a sequential one-report user-level \(\eps\)-LDP protocol as a decoded estimating-equation problem, separating the privacy guarantee for the report from the conditional-mean requirement for stochastic approximation. \item We construct the \(\mathsf{CQ}_X\left(q,s\right)\) channel, where \(q\) is the slope-coordinate grid size and \(s\) is the selected block size. The channel combines public coordinate selection, support-aware stochastic quantization, randomized response, affine decoding, and Horvitz-Thompson reconstruction using the two-face support geometry of the QR estimating-equation contribution. \item We prove local privacy and conditional decoder unbiasedness for the exact QR estimating-equation contribution. For fixed finite channel designs, we establish consistency and asymptotic normality of the projected Polyak-Ruppert averaged estimator under fixed-dimensional regularity conditions, with limiting covariance reflecting both sampling variation and privacy-channel randomization. \item We provide Hessian-free inference based on the decoded private trajectory, including a self-normalized confidence procedure for prespecified scalar contrasts and divide-and-conquer and HiGrad-style studentization procedures with explicit rank and calibration conditions. \item We empirically assess privacy-accuracy-inference tradeoffs through simulations across privacy budgets, dimensions, quantile levels, and inferential targets, together with a New York City taxi-trip illustration, benchmarking the proposed channel against direct Laplace and face-exponential geometric releases and nonprivate online stochastic approximation. \end{enumerate}

We next review related work and compare the proposed framework with adjacent method classes.

\par\textbf{ Online and communication-constrained quantile regression.} Classical QR models distributional features beyond the mean \citep{koenker1978regression,koenker2005quantile}, including projection interpretations under misspecification \citep{angrist2006quantile}. Recent online QR work studies stochastic subgradient methods when records arrive sequentially and the update is directly observable \citep{shen2025onlineqr}. Distributed and communication-constrained QR addresses data split across machines or institutions with controlled communication \citep{tan2022distributedqr}. Our focus is instead the sequential one-report LDP setting, where each record is privatized before the server forms a decoded estimating-equation input by post-processing an LDP report.

\par\textbf{Local privacy mechanisms and private regression.}
The LDP literature studies the statistical cost of replacing raw records by local randomized reports \citep{duchi2013local,duchi2018minimax}. Randomized response, RAPPOR, extremal mechanisms, subset selection, Hadamard response, and related finite-channel methods provide mechanism tools for categorical and distributional tasks \citep{warner1965randomized,erlingsson2014rappor,kairouz2016extremal,ye2018optimal,acharya2019hadamard,pastore2021rr}. A separate line studies non-interactive LDP empirical risk minimization and regression, including generalized linear models and structural assumptions that reduce sample complexity \citep{wang2020nldperm,wang2023nldpglm}.

Generic numeric-vector LDP methods can be applied to an estimating-equation vector \citep{wang2019collecting,li2019numerical,asi2022optimal,asi2023fast}, but they treat the contribution as an ambient numeric object. Our construction instead uses the QR two-face support geometry and decodes a categorical privatized report to recover the estimating-equation contribution in conditional mean.

\par\textbf{Private quantiles and private QR inference.}
Private quantile estimation has a longer central-DP literature. Smooth-sensitivity and robust-statistical constructions give early DP mechanisms for medians and related quantile-type statistics \citep{nissim2007smooth,dwork2009robust,smith2011privacy}. Recent work studies exact or approximate multiple quantiles, private quantile functions, bounded-space streaming quantiles, and high or unbounded quantiles \citep{gillenwater2021quantiles,kaplan2022approxquantiles,lalanne2023many,alabi2023bounded,durfee2023unbounded,imola2025smaller}. Under local privacy, scalar quantile and CDF procedures provide inference for one-dimensional distributional targets using binary queries, adaptive protocols, or histogram/range-query primitives \citep{liu2023online,liu2024cdf,cai2025timeuniform,aamand2025lightweight,gaboardi2019mean,cormode2019range,bassily2015local,canonne2025histograms}. These methods do not require covariate-dependent QR updates.

Related one-bit quasi-MLE and federated quantile-inference methods study complementary local or distributed protocols \citep{ono2022onebit,cai2025federated}. Central and distributed DP QR methods protect estimators or optimization procedures under curator or server-side models \citep{chen2023dpquantileloss,tran2024dpqr,shen2025highdim,lu2025versatile}, while central-DP quantile algorithms address scalar or multiple quantile release \citep{gillenwater2021quantiles,kaplan2022approxquantiles,durfee2023unbounded}. This differs from the one-report local model considered here.

\par\textbf{Inference for stochastic approximation and private outputs.}
Our estimation theory uses Polyak-Ruppert averaging and standard martingale tools for stochastic approximation \citep{ruppert1988efficient,polyak1992acceleration,robbins1971supermartingale,kushner2003stochastic,hall1980martingale,bach2011nonasymptotic}. For uncertainty quantification, self-normalization, batch means, online bootstrap, divide-and-conquer ideas, and HiGrad-style hierarchical grouping are closely related \citep{shao2010selfnormalized,chen2020statistical,zhu2021batchmeans,fang2018online,banerjee2019divide,su2023higrad}. Differentially private inference methods emphasize that valid confidence statements must account for the privatized output, not only the sampling distribution of the nonprivate estimator \citep{dette2024block,xie2025online,wang2025dpbootstrap}. Here, the privacy channel enters both the covariance and the finite-sample stability of confidence procedures.

\begin{table}[t]
\caption{ Comparison with adjacent private quantile, quantile-regression, and local-privacy methods. A \(\checkmark\) means that representative work in the method class directly addresses the requirement; a \(\times\) means that the requirement is outside that line's stated target, trust model, timing model, or inferential output.}
\label{tab:related-comparison}
\centering

\setlength{\tabcolsep}{1.05pt}
\renewcommand{\arraystretch}{1.12}
\begin{tabular}{@{}>{\raggedright\arraybackslash}p{0.245\textwidth}*{9}{>{\centering\arraybackslash}p{0.069\textwidth}}@{}}
\toprule
Requirement &
\shortstack{Scalar\\LDP\\Q/CDF} &
\shortstack{CDP\\ Q} &
\shortstack{Shuf./\\dist. Q} &
\shortstack{Online\\QR} &
\shortstack{Dist.\\QR} &
\shortstack{DP\\QR} &
\shortstack{LDP\\ERM /\\GLM} &
\shortstack{Online\\private\\SGD} &
\shortstack{\textbf{This}\\\textbf{paper}} \\
\midrule
Record randomized before server analysis &
\(\checkmark\) & \(\times\) & \(\checkmark\) & \(\times\) & \(\times\) & \(\times\) & \(\checkmark\) & \(\checkmark\) & \(\checkmark\) \\
\addlinespace[0.5pt]
One data-dependent report per participant &
\(\checkmark\) & \(\times\) & \(\checkmark\) & \(\times\) & \(\times\) & \(\times\) & \(\checkmark\) & \(\checkmark\) & \(\checkmark\) \\
\addlinespace[0.5pt]
Online stochastic update path &
\(\checkmark\) & \(\times\) & \(\times\) & \(\checkmark\) & \(\times\) & \(\times\) & \(\times\) & \(\checkmark\) & \(\checkmark\) \\
\addlinespace[0.5pt]
Regression covariates &
\(\times\) & \(\times\) & \(\times\) & \(\checkmark\) & \(\checkmark\) & \(\checkmark\) & \(\checkmark\) & \(\checkmark\) & \(\checkmark\) \\
\addlinespace[0.5pt]
Quantile-regression coefficient target &
\(\times\) & \(\times\) & \(\times\) & \(\checkmark\) & \(\checkmark\) & \(\checkmark\) & \(\times\) & \(\times\) & \(\checkmark\) \\
\addlinespace[0.5pt]
CI or confidence-set inference &
\(\checkmark\) & \(\times\) & \(\times\) & \(\times\) & \(\checkmark\) & \(\checkmark\) & \(\checkmark\) & \(\checkmark\) & \(\checkmark\) \\
\addlinespace[0.5pt]
Server update uses only LDP reports &
\(\checkmark\) & \(\times\) & \(\checkmark\) & \(\times\) & \(\times\) & \(\times\) & \(\checkmark\) & \(\checkmark\) & \(\checkmark\) \\
\bottomrule
\end{tabular}
\vspace{2pt}
\begin{minipage}{0.98\textwidth}
\footnotesize
\emph{Representative work.} Scalar LDP Q/CDF includes \citet{liu2023online,liu2024cdf,cai2025timeuniform,aamand2025lightweight}. Central-DP (CDP) scalar or multiple quantile release includes \citet{nissim2007smooth,dwork2009robust,gillenwater2021quantiles,kaplan2022approxquantiles,lalanne2023many,alabi2023bounded,durfee2023unbounded,imola2025smaller}; these papers use a trusted-curator model and do not provide locally randomized QR updates. Shuffle or distributed private quantile protocols include \citet{aamand2025lightweight}. Online QR is represented by \citet{shen2025onlineqr}, and distributed QR by \citet{tan2022distributedqr}. Central or distributed DP QR includes \citet{chen2023dpquantileloss,tran2024dpqr,shen2025highdim,lu2025versatile,wang2025dpbootstrap}. LDP ERM/GLM includes \citet{wang2020nldperm,wang2023nldpglm,ono2022onebit,wang2019collecting,li2019numerical,asi2022optimal,asi2023fast}. Online private SGD inference includes \citet{xie2025online,dette2024block}. A checkmark for a method class means that representative work in the class has the row feature; it does not mean every cited paper in the class has that feature.
\end{minipage}
\end{table}

Table~\ref{tab:related-comparison} compares the proposed setting with adjacent method classes.  To our knowledge, this is the first one-report LDP framework for online QR estimation and inference based on mean-preserving decoded estimating-equation inputs rather than direct additive-noise SGD.

The remainder is organized as follows. Section~\ref{sec:problem-setup} presents the target, sequential LDP model, decoded estimating-equation framework, support geometry, and channel. Section~\ref{sec:private-sa-inference} gives estimation and inference theory; Sections~\ref{sec:empirical-study} and~\ref{sec:real-data-taxi} report simulations and the taxi illustration; Section~\ref{sec:discussion} concludes. Appendices~\ref{app:simulation-supplement} and~\ref{app:additional-experimental-summaries} report simulation implementation details and additional numerical summaries; proofs are deferred to the supplementary proof document.

\section{Methodology and Local Privacy Mechanism}
\label{sec:problem-setup}
\subsection{Preliminaries}
Let \(W_i\in\left[-1,1\right]^p\) denote the bounded non-intercept covariate vector and set
$X_i=\left(1,W_i^\top\right)^\top\in\R^{p+1}.$
The private record is \(\mathcal O_i=\left(W_i,Y_i\right)\), equivalently \(\left(X_i,Y_i\right)\) after appending the deterministic intercept.  The observations are i.i.d. copies of \(\mathcal O=\left(X,Y\right)\), and the coefficient vector \(\beta\in\R^{p+1}\) includes the intercept.  For a fixed quantile level \(\tau\in\left(0,1\right)\), write
\[
 \rho_\tau\left(u\right)=u\left\{\tau-\ind\left(u<0\right)\right\}.
\]
The target parameter is
\begin{equation}
 \bstar=\arg\min_{\beta\in\mathcal B} M\left(\beta\right),
 \qquad
 M\left(\beta\right)=\E\left\{\rho_\tau\left(Y-X^\top\beta\right)\right\},
 \label{eq:target}
\end{equation}
where \(\mathcal B\subset\R^{p+1}\) is compact and convex.  The regularity conditions in Section~\ref{sec:private-sa-inference} require uniqueness of this minimizer and the local density and nonsingularity assumptions needed for asymptotic linearization.  Define
\[
 m\left(\beta\right)=\E\left\{g_\beta\left(\mathcal O\right)\right\},
 \qquad
 g_\beta\left(\mathcal O\right)=X\left\{\ind\left(Y\le X^\top\beta\right)-\tau\right\}.
\]
We refer to \(g_\beta\left(\mathcal O\right)\) as the exact QR estimating-equation contribution.  Under local privacy this vector may be computed on the user side, but it is never transmitted to the server.  When \(Y\mid X\) has a conditional density at \(X^\top\bstar\), the Jacobian of \(m\) at the target, under the above sign convention, is
\begin{equation}
 H=\E\left\{f_{Y\mid X}\left(X^\top\bstar\mid X\right)XX^\top\right\}.
 \label{eq:H}
\end{equation}
The matrix \(H\) is the QR Hessian appearing in the asymptotic variance.  The inference procedures considered later avoid estimating \(H\) directly.

\begin{definition}[Sequential one-report local differential privacy]
\label{def:seq-ldp}
Let \(Q_i\) collect all public information fixed before user \(i\)'s record is randomized, including the current iterate, selected coordinate block, grid, privacy budget, relabeling rule, and any external randomness used by the server.  A sequentially queried one-report protocol is \(\eps\)-locally differentially private ($\eps-LDP$) for the participant's one-record contribution \(\mathcal O_i=\left(W_i,Y_i\right)\) if, conditional on every possible value of \(Q_i\), for all measurable output sets \(\mathcal U\) and all record pairs \(o,o'\),
\[
 \Prob\left\{\mathcal M_i\left(o\right)\in\mathcal U\mid Q_i\right\}
 \le
 e^\eps
 \Prob\left\{\mathcal M_i\left(o'\right)\in\mathcal U\mid Q_i\right\}.
\]
\end{definition}

This is the standard conditional formulation for LDP protocols with public or predictable sequential queries \citep{duchi2013local,dwork2014algorithmic}.  The budget \(\eps\) is per user: all design choices may depend on past LDP reports and external randomness, but must be fixed before the current record is randomized, while decoding, reconstruction, averaging, comparisons, and confidence sets are post-processing.  The one-report constraint matches streaming settings in which each record is available once.

\begin{definition}[Decoded estimating-equation channel]
\label{def:decoded-ee-channel}
Fix a public query point \(\beta\in\mathcal B\), a public channel specification \(\theta\), a local randomization rule \(\mathcal M_{\theta,\beta}\), a public decoder \(D_{\theta,\beta}\), and a record \(o=\left(x,y\right)\).  Let
\[
 \widetilde Z=\mathcal M_{\theta,\beta}\left(o\right),
 \qquad
 \gtilde_\theta\left(\beta\right)=D_{\theta,\beta}\left(\widetilde Z\right),
\]
where \(\widetilde Z\) is the LDP report.  The pair \(\left(\mathcal M_{\theta,\beta},D_{\theta,\beta}\right)\) is an unbiased \(\eps\)-LDP decoded estimating-equation channel if \(\widetilde Z\) is \(\eps\)-LDP for the participant's one-record contribution and
\[
 \E\left\{\gtilde_\theta\left(\beta\right)\mid \mathcal O=o,\beta,\theta\right\}=g_\beta\left(o\right).
\]
For coordinate-block channels, this full-vector unbiasedness averages over the record-independent block draw.  Let \(B\) be the realized public block, \(P_B\) the coordinate projection, \(\pi_j=\Pr\left(j\in B\right)\) the public coordinate inclusion probability, and \(D_B^{\rm blk}\) the selected-block decoder.  The block decoder satisfies
\[
 \E\left\{D_B^{\rm blk}\left(\widetilde Z\right)\mid \mathcal O=o,\beta,B\right\}
 =P_Bg_\beta\left(o\right).
\]
The full vector is then formed by Horvitz-Thompson reconstruction and by averaging over the data-independent block draw:
\[
 \E\left\{\gtilde_\theta\left(\beta\right)\mid \mathcal O=o,\beta\right\}=g_\beta\left(o\right).
\]
Conditioning additionally on a realized block gives coordinate mean \(\ind\left(j\in B\right)g_{\beta,j}\left(o\right)/\pi_j\); the full-vector equality above averages over the public block randomization.
\end{definition}

The online recursion uses the decoded input, not the LDP report.  The query
\(\beta_{i-1}\) is determined by the server state before the current
public block is drawn.  Section~\ref{sec:private-sa-inference} gives the
pre-block/post-block conditioning distinction; the full mean field is recovered
only after averaging over the fresh record-independent block.
With \(\Pi_{\mathcal B}\) denoting Euclidean projection onto \(\mathcal B\), the projected stochastic-approximation update is
\[
 \beta_i=\Pi_{\mathcal B}\left\{\beta_{i-1}-\eta_i\gtilde_i\left(\beta_{i-1}\right)\right\},
 \qquad
 \eta_i=\eta_0 i^{-\gamma},\quad \gamma\in\left(1/2,1\right).
\]
The decoded sequence has the form $
 \gtilde_1\left(\beta_0\right),\ \gtilde_2\left(\beta_1\right),\ldots,\gtilde_n\left(\beta_{n-1}\right), $
and the reported estimator is the Polyak average
\begin{equation}
 \bar\beta_n=n^{-1}\sum_{i=1}^n\beta_i.
 \label{eq:polyak}
\end{equation}
At a fixed \(\beta\), the privacy-channel noise is \(\gtilde_\theta\left(\beta\right)-g_\beta\left(\mathcal O\right)\), whereas the martingale estimating-equation noise in the stochastic-approximation analysis is \(\gtilde_\theta\left(\beta\right)-m\left(\beta\right)\).  At \(\bstar\), where \(m\left(\bstar\right)=0\),
\[
 \gtilde_\theta\left(\bstar\right)
 =g_{\bstar}\left(\mathcal O\right)
 +\left\{\gtilde_\theta\left(\bstar\right)-g_{\bstar}\left(\mathcal O\right)\right\},
\]
so the decoded input contains both the ordinary QR estimating-equation fluctuation and the additional channel noise induced by local randomization.


The finite-alphabet channel below exploits the exact support of a single-record QR contribution.  For \(x=\left(1,z^\top\right)^\top\) with \(z\in\left[-1,1\right]^p\), define
\[
 S_\beta\left(o\right)=\ind\left(y\le x^\top\beta\right)-\tau\in\left\{-\tau,1-\tau\right\}.
\]
Then
\[
 g_\beta\left(o\right)=xS_\beta\left(o\right),
 \qquad
 g_{\beta,0}\left(o\right)=S_\beta\left(o\right),
 \qquad
 g_{\beta,j}\left(o\right)=S_\beta\left(o\right)z_j,\quad j=1,\ldots,p .
\]
Thus the exact support of \(g_\beta\left(o\right)\) is
\[
 \mathcal G_\tau
 =
 \left\{-\tau\right\}\times\left[-\tau,\tau\right]^p
 \;\cup\;
 \left\{1-\tau\right\}\times\left[-\left(1-\tau\right),1-\tau\right]^p,
\]
with \(\ell_1\)-diameter
\begin{equation}
 \Delta_1
 =\sup_{u,v\in\mathcal G_\tau}\norm{u-v}_1
 =\max\left\{p+1,\,2p\max\left(\tau,1-\tau\right)\right\}.
 \label{eq:app-delta1}
\end{equation}
The intercept coordinate identifies the active face.  Conditional on this face, each slope coordinate lies in a symmetric interval with radius
\[
|S_\beta\left(o\right)|=
 \begin{cases}
 \tau, & S_\beta\left(o\right)=-\tau,\\
 1-\tau, & S_\beta\left(o\right)=1-\tau.
 \end{cases}
\]
Let \(r_{\max}=\max\left\{\tau,1-\tau\right\}\). If the face coordinate is available, selected slopes can be encoded on the covariate scale because \(g_j/g_0=z_j\in\left[-1,1\right]\); otherwise the channel uses the unconditional estimating-equation interval \(\left[-r_{\max},r_{\max}\right]\).

This geometry explains why several natural private releases are not well matched to online QR. Scalar LDP quantile protocols protect one-dimensional threshold information, but the QR update requires the product of a residual comparison and the covariate vector. Separately releasing covariates and residual signs would require multiple LDP reports or a product-category construction, without using the two-face support or directly providing the decoded full-vector contribution used by the online recursion. Generic numeric-vector LDP mechanisms can be applied to \(g_\beta\left(\mathcal O\right)\), but treat the update as an ambient vector. The two geometric baselines below isolate the resulting tradeoff: direct Laplace release preserves the mean but ignores support, whereas a face-restricted exponential release preserves support but generally shifts the mean.

A direct Laplace estimating-equation release sends
\begin{equation}
 Z^{\rm Lap}=g_\beta\left(\mathcal O\right)+\ell^{\rm Lap},
 \qquad
 \ell^{\rm Lap}_j\stackrel{\rm iid}{\sim}\mathrm{Laplace}\left(\Delta_1/\eps\right),
 \quad j=0,\ldots,p .
 \label{eq:app-laplace-release}
\end{equation}
Because \(\norm{g_\beta\left(o\right)-g_\beta\left(o'\right)}_1\le \Delta_1\) for all records \(o,o'\), the release in \eqref{eq:app-laplace-release} is \(\eps\)-LDP. It is unbiased under the identity decoder, but its output lies in \(\R^{p+1}\), not in \(\mathcal G_\tau\), so privacy noise is spent in infeasible QR directions. Each coordinate receives privacy-noise variance \(2\left(\Delta_1/\eps\right)^2\), a cost especially visible when \(p+1\) is moderate or \(\eps\) is small because the \(\ell_1\)-diameter in \eqref{eq:app-delta1} governs all coordinates simultaneously.

A support-preserving face-exponential mechanism draws \(Z^{\rm EM}\in\mathcal G_\tau\) from
\begin{equation}
 \Pr\left(Z^{\rm EM}\in dv\mid g_\beta\left(\mathcal O\right)\right)
 \propto
 \exp\!\left\{-\frac{\eps}{2\Delta_1}\norm{v-g_\beta\left(\mathcal O\right)}_1\right\}\mu_\tau\left(dv\right),
 \qquad v\in\mathcal G_\tau,
 \label{eq:app-face-em}
\end{equation}
where \(\mu_\tau\) is the sum of the \(p\)-dimensional Lebesgue measures on the two faces, with counting measure in the degenerate case \(p=0\).  The utility sensitivity is at most \(\Delta_1\), so \eqref{eq:app-face-em} is \(\eps\)-LDP.  However, the conditional mean of \(Z^{\rm EM}\) generally differs from \(g_\beta\left(\mathcal O\right)\), because the density is truncated on the two faces and also mixes between them.  If this release is inserted directly into stochastic approximation, the recursion follows a perturbed mean field rather than \(m\left(\beta\right)\).  Thus support preservation alone is not enough for QR estimation; the server-side update must also be correct in conditional mean.

These alternatives isolate the two failures that the proposed channel avoids. Separate private releases of covariates and residual signs do not directly produce the coupled QR contribution and would require multiple reports or a larger product category. Direct Laplace release preserves the estimating-equation mean but spends privacy noise outside the QR support, whereas the face-exponential release preserves the two-face support but generally shifts the conditional mean and hence the stochastic-approximation target. The coordinate-quantized channel instead privatizes one selected-block finite category and applies an affine randomized-response inverse, so the decoded server-side input is mean preserving after Horvitz-Thompson reconstruction, even though the final decoded vector need not lie in \(\mathcal G_\tau\). Section~\ref{sec:empirical-study} gives numerical evidence that the proposed LDP mechanism outperforms these two designs in the reported regimes for estimation error and uncertainty quantification.

\begin{figure}[H]
\centering
\includegraphics[width=0.98\textwidth]{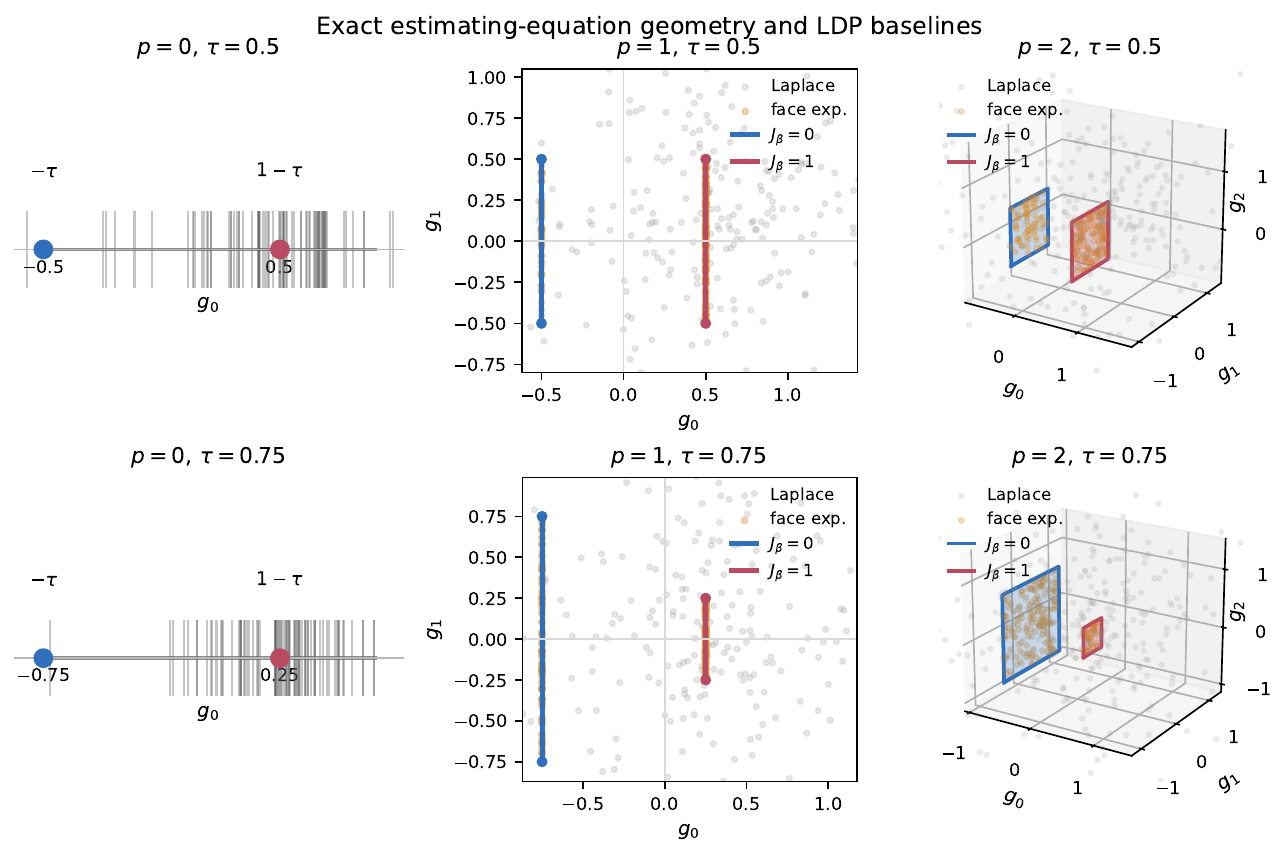}
\caption{Feasible exact local QR estimating-equation contributions and geometric baseline LDP releases for \(p=0,1,2\).  Here \(J_\beta=\ind\left(Y\le X^\top\beta\right)\), so \(J_\beta=0\) and \(J_\beta=1\) are the two faces.  Blue and red objects show the exact support.  Gray marks show direct-Laplace releases: vertical ticks on the \(g_0\) axis for \(p=0\), and points in the displayed estimating-equation coordinates for \(p\ge1\).  Orange marks show face-exponential releases for \(p\ge1\).  The panels illustrate why support awareness must be paired with mean-preserving decoding.}
\label{fig:estimating-equation-regions}
\end{figure}

Figure~\ref{fig:estimating-equation-regions} illustrates the two-face QR support for \(p=0,1,2\) and \(\tau=0.5,0.75\), together with the two geometric baselines used in the numerical study. A direct Laplace release preserves the estimating-equation mean by adding centered ambient noise but ignores the feasible set. A face-exponential release remains on the feasible support but generally does not preserve the conditional mean. The coordinate-quantized channel instead privatizes a support-aware finite category for the selected block and applies a public affine inverse, yielding a decoded update with the correct conditional mean even though the final decoded vector is not constrained to lie in \(\mathcal G_\tau\).

\subsection{Coordinate-Quantized Mechanism \(\mathsf{CQ}_X\left(q,s\right)\)}

We now specify the channel \(\mathsf{CQ}_X\left(q,s\right)\). Estimating-equation coordinates are indexed by \(0,1,\ldots,p\), with coordinate \(0\) the intercept and active face, and coordinates \(1,\ldots,p\) the slopes associated with \(W\). The integer \(q\in\left\{2,3,\ldots\right\}\) is the number of slope-grid levels, and \(s\in\left\{1,\ldots,p+1\right\}\) is the number of selected estimating-equation coordinates in one report. Let \(B\subset\left\{0,\ldots,p\right\}\) be the public selected block, with \(|B|=s\). The finite-\(q\) channel is used for estimation and inference; \(q=\infty\) is reserved for nonprivate limiting references below.

The selected block is sampled uniformly without replacement from the \(p+1\) estimating-equation coordinates and is independent of the current record.  Therefore
\[
 \pi_j=\Pr\left(j\in B\right)=\frac{s}{p+1},\qquad j=0,\ldots,p.
\]
For \(p=0\), the selected block is necessarily \(B=\left\{0\right\}\) and \(\pi_0=1\).  The Horvitz-Thompson factor \(1/\pi_j\) is applied only to selected coordinates; unselected coordinates are filled with zero.

Fix a realized block \(B\) and write \(g=g_\beta\left(o\right)\).  Stochastic rounding first maps the selected coordinates to a finite latent category without introducing discretization bias.  If \(0\in B\), the intercept coordinate is the two-point face variable
$
 g_0\in\left\{-\tau,1-\tau\right\}.$
For each selected slope coordinate \(j\in B\setminus\left\{0\right\}\), define $
 u_j= g_j/g_0=z_j\in\left[-1,1\right],$
which is well defined because \(\tau\in\left(0,1\right)\).  Let
\[
 \mathcal A_q=\left\{-1+\frac{2\ell}{q-1}:\ell=0,\ldots,q-1\right\}.
\]
Adjacent-grid stochastic rounding produces \(\zeta_j\in\mathcal A_q\) with $
 \E\left(\zeta_j\mid g_j,g_0\right)=u_j.$
The selected-block representative uses \(g_0\zeta_j\) on the estimating-equation scale and is therefore unbiased for \(g_j\).

If \(0\notin B\), the report contains no face label.  Each selected slope coordinate is rounded directly on the public grid
\[
 \mathcal A_q\left(r_{\max}\right)=
 \left\{-r_{\max}+\frac{2r_{\max}\ell}{q-1}:\ell=0,\ldots,q-1\right\},
\]
again using adjacent-grid stochastic rounding with conditional mean equal to the exact selected coordinate.  In both cases, the quantization step produces a latent category \(\Lambda_B\) and a selected-block representative \(v_B\left(\Lambda_B\right)\) satisfying
\begin{equation}
 \E\left\{v_B\left(\Lambda_B\right)\mid g,B\right\}=P_Bg.
 \label{eq:block-rounding-unbiasedness}
\end{equation}

The entire privacy budget is spent on one categorical release for the selected block.  Let $
 \iota_B=\ind\left(0\in B\right),
 s_B^{\rm slope}=|B|-\iota_B$ and the block alphabet size is
$
 K_B=2^{\iota_B}q^{s_B^{\rm slope}}.
$
For this fixed block, write the latent category set as \(\mathcal K_B=\left\{1,\ldots,K_B\right\}\).  The public map \(v_B:\mathcal K_B\to\R^{|B|}\) assigns to each category its selected-block representative on the estimating-equation scale.  The user never releases \(\Lambda_B\) or \(v_B\left(\Lambda_B\right)\).

The privatized report is the randomized-response output index \(\widetilde\Lambda\):
\begin{equation}
 \Pr\left(\widetilde\Lambda=\lambda\mid \Lambda_B=\lambda\right)=\frac{e^\eps}{e^\eps+K_B-1},
 \qquad
 \Pr\left(\widetilde\Lambda=\lambda'\mid \Lambda_B=\lambda\right)=\frac{1}{e^\eps+K_B-1},\quad \lambda'\ne\lambda.
\label{eq:block-randomized-response}
\end{equation}
Define the randomized-response contraction factor
\begin{equation}
 \kappa_{\rm rr}\left(K_B,\eps\right)=\frac{e^\eps-1}{e^\eps+K_B-1}
\label{eq:block-rr-contraction}
\end{equation}
and the uniform selected-grid mean
\begin{equation}
 \bar v_B=K_B^{-1}\sum_{\lambda\in\mathcal K_B}v_B\left(\lambda\right).
\label{eq:block-grid-mean}
\end{equation}
With the public relabeling suppressed in the notation,
\[
 \E\left\{v_B\left(\widetilde\Lambda\right)\mid \Lambda_B=\lambda,B\right\}
 =
 \bar v_B+
 \kappa_{\rm rr}\left(K_B,\eps\right)\left\{v_B\left(\lambda\right)-\bar v_B\right\}.
\]
The affine block decoder inverts this contraction:
\begin{equation}
 D_B^{\rm blk}\left(\widetilde\Lambda\right)
 =
 \bar v_B+
 \frac{v_B\left(\widetilde\Lambda\right)-\bar v_B}{\kappa_{\rm rr}\left(K_B,\eps\right)}.
\label{eq:block-affine-decoder}
\end{equation}
The full decoded estimating-equation vector is
\begin{equation}
 \gtilde_j=
 \begin{cases}
 D_{B,j}^{\rm blk}\left(\widetilde\Lambda\right)/\pi_j, & j\in B,\\
 0, & j\notin B.
 \end{cases}
\label{eq:block-ht-reconstruction}
\end{equation}
The decoded range may extend beyond the exact two-face support because the affine decoder corrects randomized-response shrinkage and the Horvitz-Thompson step inflates selected coordinates.  This has no privacy consequence, since decoding and reconstruction are post-processing of the privatized report.  It does affect the variance through \(K_B\), \(\kappa_{\rm rr}\left(K_B,\eps\right)\), and \(\pi_j\). The full procedure is summarized in Algorithm~\ref{alg:private-online-qr}.

\begin{algorithm}[H]
\caption{Locally private online quantile-regression ASGD}
\label{alg:private-online-qr}

\footnotesize
\begin{enumerate}[leftmargin=*,itemsep=1pt,topsep=1pt]
 \item Given \(\tau\), \(\left(q,s,\eps\right)\), \(\mathcal B\), \(\beta_0\), \(\eta_i=\eta_0i^{-\gamma}\) with \(\gamma\in\left(1/2,1\right)\), stream length \(n\), and reporting set \(\mathcal T\subset\left\{1,\ldots,n\right\}\), set \(\bar\beta_0=\beta_0\).
 \item For \(i=1,\ldots,n\):
 \begin{enumerate}[leftmargin=*,label=(\alph*),itemsep=1pt,topsep=1pt]
 \item Draw \(B_i\subset\left\{0,\ldots,p\right\}\) uniformly without replacement with \(|B_i|=s\), independently of \(\mathcal O_i=\left(X_i,Y_i\right)\), set \(\pi_j=s/\left(p+1\right)\), and send \(\left(\beta_{i-1},B_i,q,\eps,\tau\right)\) and the public relabeling convention to user \(i\).
 \item User \(i\) forms \(g_i=g_{\beta_{i-1}}\left(\mathcal O_i\right)\) and applies the block rounding construction above to obtain \(\Lambda_{B_i}\), so that \eqref{eq:block-rounding-unbiasedness} holds with \(B=B_i\) and \(g=g_i\).
 \item User \(i\) sends only the privatized report \(\widetilde\Lambda_i\), generated from \(\Lambda_{B_i}\) by the randomized-response law \eqref{eq:block-randomized-response}.
 \item The server evaluates \(\kappa_{\rm rr}\left(K_{B_i},\eps\right)\) from \eqref{eq:block-rr-contraction}, \(\bar v_{B_i}\) from \eqref{eq:block-grid-mean}, decodes by \eqref{eq:block-affine-decoder}, and forms \(\gtilde_i\left(\beta_{i-1}\right)\) by the reconstruction rule \eqref{eq:block-ht-reconstruction}.
 \item Update
 \[
  \beta_i
  =
  \Pi_{\mathcal B}\left\{\beta_{i-1}-\eta_i\gtilde_i\left(\beta_{i-1}\right)\right\},
  \qquad
  \bar\beta_i=(i-1)i^{-1}\bar\beta_{i-1}+i^{-1}\beta_i .
 \]
 Store \(\bar\beta_i\) if \(i\in\mathcal T\), and store trajectory values needed for post-processing inference.
 \end{enumerate}
 \item Return \(\bar\beta_n\), \(\left\{\bar\beta_t:t\in\mathcal T\right\}\), and the stored decoded trajectory.
\end{enumerate}

\end{algorithm}

The next proposition collects the privacy, unbiasedness, finite-range, and local-stability properties of \(\mathsf{CQ}_X\left(q,s\right)\) used by the online analysis.

\begin{proposition}[Privacy, unbiasedness, and finite range of \(\mathsf{CQ}_X\left(q,s\right)\)]
\label{prop:unbiased-decoded-input}
For any finite \(q<\infty\), any \(1\le s\le p+1\), and any \(\eps>0\), the channel \(\mathsf{CQ}_X\left(q,s\right)\) satisfies the following properties.
\begin{enumerate}[leftmargin=*]
 \item Conditional on the public selected block, the randomized-response output \(\widetilde\Lambda\) is \(\eps\)-locally differentially private for the participant's one-record contribution.  Mixing over data-independent selected blocks preserves the same guarantee.
 \item The selected-block decoder and full Horvitz-Thompson reconstruction satisfy
 \[
  \E\left\{D_B^{\rm blk}\left(\widetilde\Lambda\right)\mid g,B\right\}=P_Bg,
  \qquad
  \E\left(\gtilde\mid g\right)=g,
 \]
 where the second expectation also averages over the data-independent public block draw.
 \item For every fixed finite design \(\left(q,s,\eps\right)\), the decoded vector is uniformly bounded over all records, query points, selected blocks, and possible randomized-response outputs:
 \[
  \sup_{\beta\in\mathcal B}\sup_o\sup_{\widetilde\Lambda}
  \norm{\gtilde}_{2}
  \le C_{\rm CQ}\left(p,\tau,q,s,\eps\right)<\infty .
 \]
 \item Under a common implementation of the public block draw, stochastic rounding, and randomized response, two decoded inputs queried at \(\beta\) and \(\beta'\) are identical for a fixed record whenever
 \[
  \ind\left(y\le x^\top\beta\right)=\ind\left(y\le x^\top\beta'\right).
 \]
 Hence local changes in the decoded input occur only on the usual QR slab between the two hyperplanes.
\end{enumerate}
\end{proposition}

\label{sec:privacy-budget-limits}

The parameters \(q\), \(s\), and \(\eps\) control different parts of the privacy-accuracy tradeoff.  The grid size \(q\) sets the resolution of the selected slope coordinates: larger \(q\) reduces stochastic-rounding variability but increases the alphabet size $ K_B=2^{\ind\left(0\in B\right)}q^{|B|-\ind\left(0\in B\right)}.$ For fixed \(\eps\), larger \(K_B\) decreases $ \kappa_{\rm rr}\left(K_B,\eps\right)$ in (\ref{eq:block-rr-contraction}), so the affine decoder amplifies centered randomized-response noise by \(1/\kappa_{\rm rr}\left(K_B,\eps\right)\).  Thus \(q\) is not monotone for accuracy at finite privacy budget: it trades quantization error against randomized-response inversion noise.

The block size \(s\) determines how many estimating-equation coordinates are queried in one local report.  Larger \(s\) reduces Horvitz-Thompson inflation because \(\pi_j=s/(p+1)\) increases, and \(s=p+1\) removes coordinate subsampling.  At the same time, larger \(s\) usually increases \(K_B\), so it may increase randomized-response noise when \(\eps\) is fixed.  The design therefore spends the entire privacy budget on one selected-block category rather than splitting \(\eps\) coordinate by coordinate.

The privacy budget \(\eps\) acts only through the randomized-response channel.  For fixed \(K_B\),
\[
 \kappa_{\rm rr}\left(K_B,\eps\right)\uparrow 1
 \quad\text{as}\quad
 \eps\to\infty,
 \qquad
 \kappa_{\rm rr}\left(K_B,\eps\right)\downarrow 0
 \quad\text{as}\quad
 \eps\downarrow0.
\]
At small \(\eps\), \(\kappa_{\rm rr}\left(K_B,\eps\right)\approx \eps/K_B\), so affine decoding can greatly inflate server-side variance.  At large \(\eps\), randomized-response noise vanishes, leaving only finite-grid and coordinate-subsampling losses.

Several special cases connect \(\mathsf{CQ}_X\left(q,s\right)\) to familiar procedures.
\begin{enumerate}[leftmargin=*]
 \item \textbf{Scalar quantile case.}  If \(p=0\), then \(B=\left\{0\right\}\), \(K_B=2\), and \(q\) is irrelevant.  The mechanism reduces to binary randomized response applied to $
  S_\beta\left(Y\right)=\ind\left(Y\le \beta\right)-\tau. $
 The resulting recursion is the scalar online LDP quantile stochastic-approximation scheme, with affine decoding of the randomized binary response \citep{liu2023online}.
 \item \textbf{Coarsest finite member.}  When \(q=2\), adjacent-grid stochastic rounding reduces each selected slope to one of the two endpoints of its public interval.  This is the coarsest stochastic-rounding member of the family, analogous to unbiased low-precision or quantized-gradient schemes \citep{gupta2015limitedprecision,alistarh2017qsgd}.  With \(s=1\), each user reports one randomized selected-coordinate category.  It has low communication and small alphabet size, but it uses only one coordinate from each participant contribution and therefore relies heavily on Horvitz-Thompson reconstruction.
 \item \textbf{All-coordinate finite-grid release.}  When \(s=p+1\) and \(q<\infty\), all estimating-equation coordinates are selected and \(\pi_j=1\).  The channel becomes a finite-grid randomized-response release of the whole QR contribution.  Since the intercept is always included, the selected slopes are quantized on the covariate scale and then mapped back to the estimating-equation scale.
 \item \textbf{No randomized-response noise.}  If \(\eps=\infty\) while \(q<\infty\), then \(\kappa_{\rm rr}\left(K_B,\eps\right)=1\) and the randomized-response step becomes the identity.  This regime corresponds to finite-precision stochastic approximation with quantized arithmetic or quantized gradients, as in practical low-precision CPU/GPU implementations and communication-efficient SGD \citep{gupta2015limitedprecision,alistarh2017qsgd}.  The procedure is no longer private, but finite stochastic quantization and, if \(s<p+1\), coordinate subsampling remain.  The decoded input is still conditionally unbiased for \(g_\beta\left(\mathcal O\right)\).
 \item \textbf{Private continuous-response releases.}  If \(q=\infty\) while \(\eps<\infty\), the finite-alphabet randomized-response construction is replaced by a continuous LDP release of the bounded estimating-equation contribution, for example through Laplace-type noise calibrated to the contribution's public sensitivity.  Such mechanisms are useful baselines, but they are outside the finite-channel theory developed here.
 \item \textbf{Nonprivate online QR reference.}  The simultaneous limit
 $
  q=\infty, s=p+1, \eps=\infty
 $
 removes stochastic quantization, coordinate subsampling, and randomized-response noise.  In this limit, $
  \gtilde_i\left(\beta_{i-1}\right)=g_{\beta_{i-1}}\left(\mathcal O_i\right),$
 and the update reduces to the ordinary nonprivate projected averaged stochastic-approximation algorithm for QR, as in nonprivate online QR and Polyak-Ruppert averaging \citep{shen2025onlineqr,ruppert1988efficient,polyak1992acceleration}.
\end{enumerate}

Thus \(\mathsf{CQ}_X\left(q,s\right)\) is a coordinate-quantized template for bounded estimating-equation contributions beyond QR; below we focus mainly on \(q<\infty\) and \(\eps<\infty\).

\section{Asymptotic Properties}
\label{sec:private-sa-inference}
\subsection{Assumptions and Main Results}
We now study the decoded online trajectory produced by the finite channel in Section~\ref{sec:problem-setup}. Recall that the privatized report of user \(i\) is denoted by \(\widetilde Z_i\); for the channel \(\mathsf{CQ}_X\left(q,s\right)\), this is the randomized category \(\widetilde\Lambda_i\) in Algorithm~\ref{alg:private-online-qr}.  The decoded vector that the server would obtain at query point \(\beta\) is denoted by \(\gtilde_i\left(\beta\right)\), and the actually observed decoded input is \(\gtilde_i\left(\beta_{i-1}\right)\).

Let \(\mathcal P_t\) collect the public server randomness at completed step
\(t\), including \(B_t\) and any public relabeling.  Fixed design quantities are
deterministic, or equivalently included in \(\mathcal H_0\).  For the analysis,
define the pre-block server history
\[
 \mathcal H_{i-1}
 =
 \sigma\left\{\beta_0,\mathcal P_1,\widetilde Z_1,\ldots,
        \mathcal P_{i-1},\widetilde Z_{i-1}\right\}.
\]
This analysis filtration is not an online-storage requirement: the
implementation may keep only its recursive state \(S_{i-1}\), such as
\(\beta_{i-1}\), \(\bar\beta_{i-1}\), online normalizers, or trajectory
summaries; this state is \(\mathcal H_{i-1}\)-measurable.  The iterate, step
size, grid, privacy budget, coordinate-selection law, and relabeling rules are
also \(\mathcal H_{i-1}\)-measurable.  The public block \(B_i\) is drawn after
\(\mathcal H_{i-1}\), independently of \(\mathcal H_{i-1}\), the current record
\(\mathcal O_i=\left(X_i,Y_i\right)\), and the current local-randomizer
auxiliary randomness.  Write
\[
 \mathcal F_{i-1}^{+}=\mathcal H_{i-1}\vee\sigma\left(B_i\right)
\]
for the enlarged public state after the block draw.
Conditional on \(\mathcal F_{i-1}^{+}\), the local randomizer
\(\mathcal O_i\mapsto \widetilde Z_i\) is \(\eps\)-LDP for the participant's
one-record contribution.  All decoded estimators, averages, normalizers,
confidence intervals, and ellipsoids below are post-processing of the
server-side transcript of LDP reports.

For stochastic approximation, martingale differences are conditioned on the pre-block history \(\mathcal H_{i-1}\), not the enlarged post-block state \(\mathcal F_{i-1}^{+}\).
Define the Horvitz-Thompson block operator
\[
 \left\{\mathcal R_B v\right\}_j
 =
 \frac{\ind\left(j\in B\right)}{\pi_j}v_j,
 \qquad j=0,\ldots,p .
\]
For every \(\mathcal H_{i-1}\)-measurable query point \(\beta\), the selected-block reconstruction has the conditional means
\[
 \E\left\{\gtilde_i\left(\beta\right)\mid \mathcal O_i,\mathcal F_{i-1}^{+}\right\}
 =
 \mathcal R_{B_i}g_\beta\left(\mathcal O_i\right),
 \qquad
 \E\left\{\gtilde_i\left(\beta\right)\mid \mathcal F_{i-1}^{+}\right\}
 =
 \mathcal R_{B_i}m\left(\beta\right).
\]
Thus the post-block target is the Horvitz-Thompson selected vector, and
\(\mathcal R_{B_i}m\left(\beta\right)\) is generally not the full mean field
\(m\left(\beta\right)\).  Averaging over the current public block gives
\begin{equation}
 \E\left\{\gtilde_i\left(\beta\right)\mid \mathcal O_i,\mathcal H_{i-1}\right\}
 =
 g_\beta\left(\mathcal O_i\right),
 \qquad
 \E\left\{\gtilde_i\left(\beta\right)\mid \mathcal H_{i-1}\right\}
 =
 m\left(\beta\right).
 \label{eq:history-unbiasedness}
\end{equation}
Consequently, for any predictable query point \(\beta\),
$\xi_i\left(\beta\right)=\gtilde_i\left(\beta\right)-m\left(\beta\right)$
is a martingale-difference input relative to \(\mathcal H_{i-1}\), but not
relative to the realized current block.
The estimator is \eqref{eq:polyak}; public scalar normalizations used in
implementations are absorbed into the step-size constant.


The assumptions below separate the standard QR and stochastic-approximation requirements from the finite-channel property proved in Section~\ref{sec:problem-setup}.

\begin{condition}[Population QR regularity]
\label{cond:population-qr}
The observations \(\mathcal O_i=\left(X_i,Y_i\right)\) are i.i.d.; \(p+1\) is fixed; \(X=\left(1,W^\top\right)^\top\) with \(W\in\left[-1,1\right]^p\); and \(\mathcal B\subset\R^{p+1}\) is compact and convex with \(\bstar\) in its interior.  The risk \(M\left(\beta\right)=\E\left\{\rho_\tau\left(Y-X^\top\beta\right)\right\}\) is finite on \(\mathcal B\), and \(\bstar\) is the unique minimizer.  Moreover, for every \(\delta>0\),
\[
 \inf_{\beta\in\mathcal B:\|\beta-\bstar\|\ge\delta}
 \left(\beta-\bstar\right)^\top m\left(\beta\right)>0.
\]
In a neighborhood of \(X^\top\bstar\), the conditional distribution \(Y\mid X\) has a density \(f_{Y\mid X}\left(\cdot\mid X\right)\) that is uniformly bounded and uniformly locally Lipschitz in its first argument, and the Hessian \(H\) in \eqref{eq:H} is positive definite.
\end{condition}

\begin{condition}[Fixed finite decoded channel]
\label{cond:public-finite-channel}
The LDP channel is a fixed public \(\mathsf{CQ}_X\left(q,s\right)\) design with \(q<\infty\), \(1\le s\le p+1\), and fixed \(\eps>0\).  For predictable queries, \eqref{eq:history-unbiasedness} holds.  In addition, the decoded range is uniformly bounded and, under a common coupling of the public channel randomness, there is a constant \(C<\infty\) such that, for \(\beta,\beta'\) in a neighborhood of \(\bstar\),
\[
 \E\left\{\|\gtilde_i\left(\beta\right)-\gtilde_i\left(\beta'\right)\|^2\right\}
 \le C\|\beta-\beta'\|.
\]
\end{condition}

\begin{condition}[ASGD schedule]
\label{cond:asgd-schedule}
The initial value \(\beta_0\in\mathcal B\) is public.  The step sizes are public and satisfy $
 \eta_i=\eta_0i^{-\gamma},
 \eta_0>0,
 \gamma\in\left(1/2,1\right).$
\end{condition}

\begin{condition}[Rank conditions for inference]
\label{cond:inference-rank}
Let
\[
 \Omega_\eps=\Var\left\{\gtilde_i\left(\bstar\right)\right\},
 \qquad
 \Sigma_\eps=H^{-1}\Omega_\eps H^{-1}.
\]
A scalar contrast \(a^\top\bstar\) is reported only when \(a^\top\Sigma_\eps a>0\).  A full-vector ellipsoid is reported only when the corresponding limiting covariance or empirical self-normalizer is nonsingular.  Group-based Hotelling sets require more group centers than the target dimension and a full-rank empirical group covariance.
\end{condition}

Condition~\ref{cond:population-qr} is the usual fixed-dimensional QR regularity framework: bounded covariates, an interior identified target, local smoothness of the conditional density near the target hyperplane, and a nonsingular QR Hessian \citep{koenker1978regression,koenker2005quantile,angrist2006quantile,chernozhukov2011inference}.  Condition~\ref{cond:public-finite-channel} is not an additional modeling assumption: for every fixed finite \(\mathsf{CQ}_X\left(q,s\right)\) channel, Proposition~\ref{prop:unbiased-decoded-input} gives LDP, unbiased decoding, and bounded range, while the displayed local stability follows from the finite decoded range and the QR slab probability bound implied by the density condition.  Condition~\ref{cond:asgd-schedule} is the standard Polyak-Ruppert averaging schedule \citep{ruppert1988efficient,polyak1992acceleration,kushner2003stochastic,bottou2018optimization,gadat2023optimal}.  Condition~\ref{cond:inference-rank} only excludes degenerate contrasts and singular covariance estimates, as in ordinary studentized and Hotelling inference \citep{shao2010selfnormalized,shao2015self,su2023higrad}.

Under Condition~\ref{cond:population-qr}, the population mean field has the local expansion
\[
 m\left(\beta\right)=H\left(\beta-\bstar\right)+O\left(\|\beta-\bstar\|^2\right),
 \qquad \beta\to\bstar.
\]
Together with \eqref{eq:history-unbiasedness}, this yields a standard averaged stochastic-approximation problem with bounded martingale noise; the technical verification is deferred to the supplementary proof document.  The asymptotic properties are as follows.


\begin{theorem}[Consistency]
\label{thm:consistency}
Suppose Conditions~\ref{cond:population-qr}-\ref{cond:asgd-schedule} hold for a fixed finite \(\mathsf{CQ}_X\left(q,s\right)\) channel with \(q<\infty\), \(1\le s\le p+1\), and \(\eps>0\).  Then the recursion \eqref{eq:polyak} satisfies
\[
 \beta_i\to\bstar,
 \qquad
 \bar\beta_n\to\bstar,
\]
in probability.
\end{theorem}

\begin{theorem}[Polyak-Ruppert representation and CLT]
\label{thm:an}
Suppose the assumptions of Theorem~\ref{thm:consistency} hold.  Define
$
 \xi_i^\star=\gtilde_i\left(\bstar\right)-m\left(\bstar\right)=\gtilde_i\left(\bstar\right),
 \Omega_\eps=\Var\left(\xi_i^\star\right).$
Then
\begin{equation}
 \bar\beta_n-\bstar
 =
 -H^{-1}\frac1n\sum_{i=1}^n\xi_i^\star
 +o_p\left(n^{-1/2}\right),
 \label{eq:linrep}
\end{equation}
and hence
\begin{equation}
 \sqrt n\left(\bar\beta_n-\bstar\right)
 \Rightarrow
 N\left(0,H^{-1}\Omega_\eps H^{-1}\right).
 \label{eq:an}
\end{equation}
The variance \(\Omega_\eps\) is taken over the observation, the public block draw, the stochastic quantization, and the randomized-response step.
\end{theorem}

The matrix \(\Omega_\eps\) combines ordinary QR estimating-equation variability with the additional variability from local randomization.  Thus local privacy changes the estimator covariance, but not the first-order target, because the decoded input is conditionally mean preserving.

\subsection{Online Inference Procedures}
\label{subsec:inference-toolkit}

The CLT in Theorem~\ref{thm:an} involves the QR Hessian \(H\).  Estimating \(H\) under LDP would require additional locally private density or second-order information.  We therefore use three Hessian-free procedures that studentize the decoded private trajectory itself: trajectory self-normalization (SN), divide-and-conquer group studentization (DC), and hierarchical group studentization (HiGrad).  The inferential target is fixed before data collection: a scalar contrast \(a^\top\bstar\), a coordinate \(e_j^\top\bstar\), a vector contrast \(A\bstar\), or the full coefficient vector \(\bstar\).

\par \textbf{SN:}
For \(1\le \ell\le n\), write $
 \bar\beta_\ell=\ell^{-1}\sum_{i=1}^\ell\beta_i.$
At \(t=0\), define the process below to be zero.

\begin{theorem}[Self-normalized inference and functional CLT]
\label{thm:sn}
Suppose Conditions~\ref{cond:population-qr}-\ref{cond:inference-rank} hold for a fixed finite \(\mathsf{CQ}_X\left(q,s\right)\) channel.  Let \(\mathbb W_{p+1}\) be standard \((p+1)\)-dimensional Brownian motion.  Then, in \(D\left[0,1\right]^{p+1}\),
\[
 \left\{
 \frac{\lfloor nt\rfloor}{\sqrt n}
 \left(\bar\beta_{\lfloor nt\rfloor}-\bstar\right):0\le t\le1
 \right\}
 \Rightarrow
 -H^{-1}\Omega_\eps^{1/2}\mathbb W_{p+1}\left(t\right).
\]
Define the self-normalizer
\begin{equation}
 \widehat V_n^{\rm SN}
 =
 \frac1{n^2}\sum_{\ell=1}^n
 \ell^2\left(\bar\beta_\ell-\bar\beta_n\right)\left(\bar\beta_\ell-\bar\beta_n\right)^\top.
 \label{eq:Vhat}
\end{equation}
For any fixed \(a\in\R^{p+1}\) with \(a^\top\Sigma_\eps a>0\), let \(\mathbb W\) be standard one-dimensional Brownian motion and set
\[
 \mathcal T_{\rm SN}
 =
 \frac{\mathbb W\left(1\right)}
 {\left[\int_0^1\left\{\mathbb W\left(t\right)-t\mathbb W\left(1\right)\right\}^2\,dt\right]^{1/2}}.
\]
Then
\begin{equation}
 T_n^{\rm SN}\left(a\right)
 =
 \frac{\sqrt n\,a^\top\left(\bar\beta_n-\bstar\right)}
 {\left(a^\top\widehat V_n^{\rm SN}a\right)^{1/2}}
 \Rightarrow
 \mathcal T_{\rm SN}.
 \label{eq:sn-stat}
\end{equation}
If \(c_{\rm SN}\left(1-\alpha_{\rm ci}\right)\) denotes the \(\left(1-\alpha_{\rm ci}\right)\)-quantile of \(|\mathcal T_{\rm SN}|\), then
\begin{equation}
 C_{a,1-\alpha_{\rm ci}}^{\rm SN}
 =
 \left[
 a^\top\bar\beta_n
 \pm
 c_{\rm SN}\left(1-\alpha_{\rm ci}\right)
 \sqrt{\frac{a^\top\widehat V_n^{\rm SN}a}{n}}
 \right]
 \label{eq:sn-scalar-ci}
\end{equation}
has asymptotic coverage \(1-\alpha_{\rm ci}\) for \(a^\top\bstar\).

For full-vector inference, let $
 \mathbb B_{p+1}\left(t\right)=\mathbb W_{p+1}\left(t\right)-t\mathbb W_{p+1}\left(1\right)$
be a \((p+1)\)-dimensional Brownian bridge and define
\[
 \mathcal Q_{{\rm SN},p+1}
 =
 \mathbb W_{p+1}\left(1\right)^\top
 \left[\int_0^1\mathbb B_{p+1}\left(t\right)\mathbb B_{p+1}\left(t\right)^\top\,dt\right]^{-1}
 \mathbb W_{p+1}\left(1\right).
\]
If \(\widehat V_n^{\rm SN}\) is nonsingular with probability tending to one, then
\[
 n\left(\bar\beta_n-\bstar\right)^\top
 \left(\widehat V_n^{\rm SN}\right)^{-1}
 \left(\bar\beta_n-\bstar\right)
 \Rightarrow
 \mathcal Q_{{\rm SN},p+1}.
\]
Thus, with \(c_{{\rm SN},p+1}\left(1-\alpha_{\rm ci}\right)\) denoting the corresponding quantile, the ellipsoid
\begin{equation}
 C_{1-\alpha_{\rm ci}}^{\rm SN}
 =
 \left\{\beta\in\R^{p+1}:
 n\left(\bar\beta_n-\beta\right)^\top
 \left(\widehat V_n^{\rm SN}\right)^{-1}
 \left(\bar\beta_n-\beta\right)
 \le c_{{\rm SN},p+1}\left(1-\alpha_{\rm ci}\right)
 \right\}
 \label{eq:sn-ellipsoid}
\end{equation}
has asymptotic coverage \(1-\alpha_{\rm ci}\) when the rank condition holds.
\end{theorem}

SN uses a single decoded ASGD trajectory and estimates scale from the time variation of its own Polyak averages.  The limiting critical values are universal Brownian-bridge functionals and can be simulated once offline; no estimate of \(H\) or \(\Omega_\eps\) is required \citep{shao2010selfnormalized,shao2015self,fang2018online,Lee2022}.

\par \textbf{DC:}
DC partitions the stream into public disjoint groups before data collection.  Each user belongs to exactly one group, so privacy remains the same one-report user-level \(\eps\)-LDP guarantee.  Let the number of groups be \(R_{\rm DC}\), and let the group sizes \(n_g^{\rm DC}\) differ by at most one.  Run Algorithm~\ref{alg:private-online-qr} independently on each group and let $
 \widehat\beta_1^{\rm DC},\ldots,\widehat\beta_{R_{\rm DC}}^{\rm DC} $
be the terminal group Polyak averages.  Define
\begin{equation}
 \widehat\beta^{\rm DC}
 =
 \frac1{R_{\rm DC}}\sum_{g=1}^{R_{\rm DC}}\widehat\beta_g^{\rm DC},
 \qquad
 \widehat\Sigma_\beta^{\rm DC}
 =
 \frac1{R_{\rm DC}-1}
 \sum_{g=1}^{R_{\rm DC}}
 \left(\widehat\beta_g^{\rm DC}-\widehat\beta^{\rm DC}\right)
 \left(\widehat\beta_g^{\rm DC}-\widehat\beta^{\rm DC}\right)^\top.
 \label{eq:app-dc-cov}
\end{equation}

For a contrast matrix \(A\in\R^{k_A\times(p+1)}\), set $
 \widehat\psi_g^{\rm DC}=A\widehat\beta_g^{\rm DC},
 \widehat\psi^{\rm DC}
 =
 R_{\rm DC}^{-1}\sum_{g=1}^{R_{\rm DC}}\widehat\psi_g^{\rm DC},$
and let \(\widehat\Sigma_A^{\rm DC}\) be the empirical covariance matrix of
\(\widehat\psi_1^{\rm DC},\ldots,\widehat\psi_{R_{\rm DC}}^{\rm DC}\).  When \(R_{\rm DC}>k_A\) and \(\widehat\Sigma_A^{\rm DC}\) is full rank, the DC Hotelling set for \(\psi=A\bstar\) is
\begin{equation}
 C_{A,1-\alpha_{\rm ci}}^{\rm DC}
 =
 \left\{\psi\in\R^{k_A}:
 R_{\rm DC}\left(\widehat\psi^{\rm DC}-\psi\right)^\top
 \left(\widehat\Sigma_A^{\rm DC}\right)^{-1}
 \left(\widehat\psi^{\rm DC}-\psi\right)
 \le
 c_{k_A,R_{\rm DC},\alpha_{\rm ci}}^{\rm DC}
 \right\},
 \label{eq:app-dc-set}
\end{equation}
where
\begin{equation}
 c_{k_A,R_{\rm DC},\alpha_{\rm ci}}^{\rm DC}
 =
 \frac{k_A\left(R_{\rm DC}-1\right)}{R_{\rm DC}-k_A}
 F_{k_A,R_{\rm DC}-k_A}\left(1-\alpha_{\rm ci}\right).
 \label{eq:app-dc-hotelling-critical}
\end{equation}
Here \(F_{\nu_1,\nu_2}\left(1-\alpha_{\rm ci}\right)\) is the \(\left(1-\alpha_{\rm ci}\right)\)-quantile of the \(F\) distribution with \(\left(\nu_1,\nu_2\right)\) degrees of freedom.  The coefficient ellipsoid is the special case \(A=I_{p+1}\), requiring \(R_{\rm DC}>p+1\).

For a scalar contrast \(a^\top\bstar\),
\begin{equation}
 a^\top\widehat\beta^{\rm DC}
 \pm
 t_{R_{\rm DC}-1}\left(1-\alpha_{\rm ci}/2\right)
 \sqrt{\frac{\widehat\sigma_{a,\rm DC}^2}{R_{\rm DC}}},
 \qquad
 \widehat\sigma_{a,\rm DC}^2
 =
 a^\top\widehat\Sigma_\beta^{\rm DC}a,
 \label{eq:app-dc-scalar}
\end{equation}
is reported when \(\widehat\sigma_{a,\rm DC}^2>0\), where \(t_\nu\left(1-\alpha_{\rm ci}/2\right)\) denotes the corresponding Student-\(t\) quantile.

\par \textbf{HiGrad:}
HiGrad uses a public tree to produce multiple correlated group centers whose correlation pattern is known from the tree.  Let the branching vector be $
 \mathbf b_{\rm HG}=\left(b_1,\ldots,b_{L_{\rm HG}}\right),
 R_{\rm HG}=\prod_{\ell=1}^{L_{\rm HG}}b_\ell,$
where \(R_{\rm HG}\) is the number of leaves.  The tree and all segment allocations are fixed before data collection.  Let \(\nu_\ell^{\rm HG}\left(r\right)\) be the ancestor of leaf \(r\) at level \(\ell\).  Each tree segment \(\left(\ell,\nu\right)\) receives a public disjoint subset of users and produces a decoded ASGD segment estimate \(\widehat\beta_{\ell,\nu}^{\rm seg}\).  With public weights \(w_\ell^{\rm HG}\) satisfying \(\sum_{\ell=1}^{L_{\rm HG}}w_\ell^{\rm HG}=1\), define the leaf estimates
\begin{equation}
 \widehat\beta_r^{\rm HG}
 =
 \sum_{\ell=1}^{L_{\rm HG}}
 w_\ell^{\rm HG}
 \widehat\beta_{\ell,\nu_\ell^{\rm HG}\left(r\right)}^{\rm seg},
 \qquad
 r=1,\ldots,R_{\rm HG}.
 \label{eq:app-hg-leaf}
\end{equation}
Let
\[
 \widehat\beta^{\rm HG}
 =
 \frac1{R_{\rm HG}}
 \sum_{r=1}^{R_{\rm HG}}\widehat\beta_r^{\rm HG},
 \qquad
 L_\beta^{\rm HG}
 =
 \begin{pmatrix}
  \left(\widehat\beta_1^{\rm HG}\right)^\top\\
  \vdots\\
  \left(\widehat\beta_{R_{\rm HG}}^{\rm HG}\right)^\top
 \end{pmatrix}.
\]
Let \(\mathbf 1_{R_{\rm HG}}\) be the \(R_{\rm HG}\)-vector of ones.  The shared prefixes induce a public covariance template \(\Sigma_{\rm tree}\).  In the balanced implementation, one concrete template is
\begin{equation}
 \left(\Sigma_{\rm tree}\right)_{rr'}
 =
 \sum_{\ell=1}^{L_{\rm HG}}
 \frac{\left(w_\ell^{\rm HG}\right)^2}{n_{\ell,\nu_\ell^{\rm HG}\left(r\right)}}
 \ind\left\{\nu_\ell^{\rm HG}\left(r\right)=\nu_\ell^{\rm HG}\left(r'\right)\right\},
 \label{eq:app-tree-template}
\end{equation}
where \(n_{\ell,\nu}\) is the size of segment \(\left(\ell,\nu\right)\).  Let
$
 \Sigma_{\rm tree}^{+}
$
be the Moore-Penrose inverse on the centered leaf subspace, and define
\begin{equation}
 \sigma_{\rm HG}^2
 =
 R_{\rm HG}^{-2}\mathbf 1_{R_{\rm HG}}^\top
 \Sigma_{\rm tree}
 \mathbf 1_{R_{\rm HG}}.
 \label{eq:app-sigma-hg}
\end{equation}
Here \(\sigma_{\rm HG}^2\) is a public tree-design multiplier, distinct from the
data-estimated contrast variance \(\widehat\sigma_{a,\rm HG}^2\) below.

For a scalar contrast \(a\), define $ \chi_r\left(a\right)=a^\top\widehat\beta_r^{\rm HG},
 \bar\chi\left(a\right)=R_{\rm HG}^{-1}\sum_{r=1}^{R_{\rm HG}}\chi_r\left(a\right),$
and
\begin{equation}
 \widehat\sigma_{a,\rm HG}^2
 =
 \frac1{R_{\rm HG}-1}
 \left\{\chi\left(a\right)-\bar\chi\left(a\right)\mathbf 1_{R_{\rm HG}}\right\}^\top
 \Sigma_{\rm tree}^{+}
 \left\{\chi\left(a\right)-\bar\chi\left(a\right)\mathbf 1_{R_{\rm HG}}\right\}.
 \label{eq:app-hg-scalar-var}
\end{equation}
Provided \(\widehat\sigma_{a,\rm HG}^2>0\), the HiGrad scalar interval is
\begin{equation}
 a^\top\widehat\beta^{\rm HG}
 \pm
 t_{R_{\rm HG}-1}\left(1-\alpha_{\rm ci}/2\right)
 \sqrt{\widehat\sigma_{a,\rm HG}^2\sigma_{\rm HG}^2}.
 \label{eq:app-hg-scalar}
\end{equation}

For vector inference, define the centered leaf residual matrix $
 \widetilde L_\beta^{\rm HG}
 =
 L_\beta^{\rm HG}
 -
 \mathbf 1_{R_{\rm HG}}\left(\widehat\beta^{\rm HG}\right)^\top$ and the whitened covariance estimator
\begin{equation}
 \widehat\Gamma_\beta^{\rm HG}
 =
 \frac1{R_{\rm HG}-1}
 \left(\widetilde L_\beta^{\rm HG}\right)^\top
 \Sigma_{\rm tree}^{+}
 \widetilde L_\beta^{\rm HG}.
 \label{eq:app-hg-gamma}
\end{equation}
When \(R_{\rm HG}>p+1\) and \(\widehat\Gamma_\beta^{\rm HG}\) is full rank, the HiGrad coefficient ellipsoid is
\begin{equation}
 C_{1-\alpha_{\rm ci}}^{\rm HG}
 =
 \left\{\beta\in\R^{p+1}:
 \left(\sigma_{\rm HG}^2\right)^{-1}
 \left(\widehat\beta^{\rm HG}-\beta\right)^\top
 \left(\widehat\Gamma_\beta^{\rm HG}\right)^{-1}
 \left(\widehat\beta^{\rm HG}-\beta\right)
 \le
 c_{p+1,R_{\rm HG},\alpha_{\rm ci}}^{\rm HG}
 \right\},
 \label{eq:app-hg-ellipsoid}
\end{equation}
where
\begin{equation}
 c_{p+1,R_{\rm HG},\alpha_{\rm ci}}^{\rm HG}
 =
 \frac{(p+1)\left(R_{\rm HG}-1\right)}{R_{\rm HG}-(p+1)}
 F_{p+1,R_{\rm HG}-(p+1)}\left(1-\alpha_{\rm ci}\right).
 \label{eq:app-hg-hotelling-critical}
\end{equation}

The following corollary records the asymptotic calibration used by the DC and
HiGrad procedures.  The segment-wise ASGD representations are consequences of
Theorem~\ref{thm:an} applied to public substreams; they are not additional
high-level assumptions.  The remaining assumptions are design-balance and
rank conditions needed by the Student and Hotelling critical values.
\begin{corollary}
\label{cor:app-group-clt-calibration}
Suppose Conditions~\ref{cond:population-qr}--\ref{cond:inference-rank} hold for
a fixed finite \(\mathsf{CQ}_X\left(q,s\right)\) channel.  All public group and segment
recursions below are initialized at public points in \(\mathcal B\) and use the
ASGD schedule \(\eta_t=\eta_0t^{-\gamma}\) in their own local time
\(t=1,2,\ldots\).

For DC, let \(R_{\rm DC}\) be fixed, let
\(\mathcal I_g^{\rm DC}\) be public disjoint groups independent of the records,
and put \(m_n^{\rm DC}=n/R_{\rm DC}\).  Assume $
 \min_g n_g^{\rm DC}\to\infty,
 \max_{1\le g\le R_{\rm DC}}
 \left|n_g^{\rm DC}/m_n^{\rm DC}-1\right|\to0 .$
Then there exist remainders \(r_{g,n}^{\rm DC}\) such that, uniformly over the
fixed set of groups,
\[
 \widehat\beta_g^{\rm DC}-\bstar
 =
 -H^{-1}\frac1{n_g^{\rm DC}}
 \sum_{i\in\mathcal I_g^{\rm DC}}\xi_i^\star
 +r_{g,n}^{\rm DC},
 \qquad
 \max_g\sqrt{n_g^{\rm DC}}\,
 \left\|r_{g,n}^{\rm DC}\right\|\stackrel{p}{\longrightarrow}0 .
\]
In particular, for any fixed contrast matrix \(A\), $
 \max_g\sqrt{m_n^{\rm DC}}\,
 \left\|A r_{g,n}^{\rm DC}\right\|\stackrel{p}{\longrightarrow}0 .$
If \(\Sigma_A=A\Sigma_\eps A^\top\) is positive definite and
\(R_{\rm DC}>k_A\), the statistic in \eqref{eq:app-dc-set} converges to the
Hotelling law used in \eqref{eq:app-dc-hotelling-critical}, and $
 \Pr\left\{A\bstar\in C_{A,1-\alpha_{\rm ci}}^{\rm DC}\right\}
 \to 1-\alpha_{\rm ci}. $
For any scalar contrast \(a\) with \(a^\top\Sigma_\eps a>0\), the DC pivot in
\eqref{eq:app-dc-scalar} converges to \(t_{R_{\rm DC}-1}\), so the scalar
interval has asymptotic coverage \(1-\alpha_{\rm ci}\).

For HiGrad, let the branching vector and weights be fixed, let
\(\mathcal I_{\ell,\nu}^{\rm HG}\) be public disjoint segment sets independent
of the records, and assume $
 \min_{\ell,\nu}n_{\ell,\nu}\to\infty . $
Then there exist remainders \(r_{\ell,\nu,n}^{\rm HG}\) such that, uniformly
over the fixed set of tree segments,
\[
 \widehat\beta_{\ell,\nu}^{\rm seg}-\bstar
 =
 -H^{-1}\frac1{n_{\ell,\nu}}
 \sum_{i\in\mathcal I_{\ell,\nu}^{\rm HG}}\xi_i^\star
 +r_{\ell,\nu,n}^{\rm HG},
 \qquad
 \max_{\ell,\nu}\sqrt{n_{\ell,\nu}}\,
 \left\|r_{\ell,\nu,n}^{\rm HG}\right\|\stackrel{p}{\longrightarrow}0 .
\]
Assume \(\sigma_{{\rm HG},n}^2>0\) eventually and $ M_n=\Sigma_{{\rm tree},n}/\sigma_{{\rm HG},n}^2\to M,$
where \(M\) has constant row sums and is positive definite on the centered leaf
subspace.  Then the weighted leaf-scale segment remainders are automatically
negligible:
\[
 \max_{1\le r\le R_{\rm HG}}
 \frac1{\sigma_{{\rm HG},n}}
 \left\|
 \sum_{\ell=1}^{L_{\rm HG}}w_\ell^{\rm HG}
 r_{\ell,\nu_\ell^{\rm HG}(r),n}^{\rm HG}
 \right\|
 \stackrel{p}{\longrightarrow}0.
\]
Consequently, for any scalar contrast \(a\) with \(a^\top\Sigma_\eps a>0\), the
HiGrad pivot underlying \eqref{eq:app-hg-scalar} converges to
\(t_{R_{\rm HG}-1}\), and the scalar interval has asymptotic coverage
\(1-\alpha_{\rm ci}\).  If \(\Sigma_\eps\) is positive definite and
\(R_{\rm HG}>p+1\), the quadratic form in \eqref{eq:app-hg-ellipsoid}
converges to the Hotelling law used in \eqref{eq:app-hg-hotelling-critical},
and
\[
 \Pr\left\{\bstar\in C_{1-\alpha_{\rm ci}}^{\rm HG}\right\}
 \to 1-\alpha_{\rm ci}.
\]

\end{corollary}

SN, DC, and HiGrad estimate scale from decoded private outputs in different ways: SN uses temporal variation in one trajectory and is calibrated by Theorem~\ref{thm:sn}; DC uses independent public groups under the balanced-group condition in Corollary~\ref{cor:app-group-clt-calibration}; and HiGrad uses a public tree with a known shared-prefix covariance template under the stated convergence, constant-row-sum, and centered-rank conditions.  All three are post-processing procedures requiring no additional raw-data access or LDP reports, and Algorithms~\ref{alg:sn-inference}--\ref{alg:higrad-inference} summarize their implementations.

\begin{algorithm}[H]
\caption{Online LDP QR inference via SN}
\label{alg:sn-inference}
\begin{enumerate}[leftmargin=*]
 \item \textbf{Input.}  Stored decoded ASGD averages \(\left\{\bar\beta_\ell:1\le\ell\le n\right\}\), target contrast \(a\) or vector target, and nominal level \(1-\alpha_{\rm ci}\).
 \item Compute \(\widehat V_n^{\rm SN}\) from \eqref{eq:Vhat}.
 \item For a scalar contrast \(a^\top\bstar\), compute \(a^\top\widehat V_n^{\rm SN}a\).  If it is positive, report the interval \eqref{eq:sn-scalar-ci}.
 \item For a coefficient ellipsoid, check whether \(\widehat V_n^{\rm SN}\) is full rank.  If so, report \eqref{eq:sn-ellipsoid}.
 \item \textbf{Output.}  SN intervals or ellipsoids.
\end{enumerate}
\end{algorithm}

\begin{algorithm}[H]
\caption{Online LDP QR inference via DC}
\label{alg:dc-inference}
\begin{enumerate}[leftmargin=*]
 \item \textbf{Input.}  Public number of groups \(R_{\rm DC}\), channel design \(\left(q,s,\eps\right)\), ASGD settings, target \(a\) or \(A\), and nominal level \(1-\alpha_{\rm ci}\).
 \item Partition users into \(R_{\rm DC}\) disjoint public groups with nearly equal sizes.
 \item Run the decoded ASGD recursion separately on each group and compute terminal group averages \(\widehat\beta_g^{\rm DC}\).
 \item Compute \(\widehat\beta^{\rm DC}\) and \(\widehat\Sigma_\beta^{\rm DC}\) from \eqref{eq:app-dc-cov}.
 \item For scalar targets, report \eqref{eq:app-dc-scalar} when the scalar variance is positive.
 \item For vector targets, report \eqref{eq:app-dc-set} when \(R_{\rm DC}>k_A\) and \(\widehat\Sigma_A^{\rm DC}\) is full rank.
 \item \textbf{Output.}  DC intervals or Hotelling sets.
\end{enumerate}
\end{algorithm}
\begin{algorithm}[ht]

\caption{Online LDP QR inference via HiGrad}
\label{alg:higrad-inference}
\begin{enumerate}[leftmargin=*]
 \item \textbf{Input.}  Public branching vector \(\mathbf b_{\rm HG}\), public segment allocations, public weights \(w_\ell^{\rm HG}\), channel design \(\left(q,s,\eps\right)\), ASGD settings, target \(a\) or full vector, and nominal level \(1-\alpha_{\rm ci}\).
 \item For each tree segment \(\left(\ell,\nu\right)\), run decoded ASGD on the users assigned to that segment and compute \(\widehat\beta_{\ell,\nu}^{\rm seg}\).
 \item For each leaf \(r\), form \(\widehat\beta_r^{\rm HG}\) using \eqref{eq:app-hg-leaf}.
 \item Compute \(\widehat\beta^{\rm HG}\), the tree template \(\Sigma_{\rm tree}\), and \(\sigma_{\rm HG}^2\).
 \item For scalar targets, compute \(\widehat\sigma_{a,\rm HG}^2\) and report \eqref{eq:app-hg-scalar} when it is positive.
 \item For coefficient ellipsoids, compute \(\widehat\Gamma_\beta^{\rm HG}\) and report \eqref{eq:app-hg-ellipsoid} when \(R_{\rm HG}>p+1\) and \(\widehat\Gamma_\beta^{\rm HG}\) is full rank.
 \item \textbf{Output.}  HiGrad intervals or ellipsoids.
\end{enumerate}
\end{algorithm}

\subsection{Effective Information and High-Privacy-Budget Recovery}

The preceding results are fixed-\(\eps\), fixed-finite-channel statements.  For interpreting finite-sample behavior, it is useful to make the randomized-response attenuation explicit.  For binary randomized response, define
\(\kappa_{\rm bin}\left(\eps\right)=\tanh\left(\eps/2\right)\).
A one-coordinate binary decoded input has variance inflation of order
\(\kappa_{\rm bin}\left(\eps\right)^{-2}\), with
\(\kappa_{\rm bin}\left(\eps\right)^2\sim\eps^2/4\) as \(\eps\downarrow0\).  If coordinate \(j\) is selected with public probability \(\pi_j\), a first-order coordinate information index is
\begin{equation}
 n_{{\rm eff},j}
 =
 n\pi_j\kappa_{\rm bin}\left(\eps\right)^2,
 \qquad
 n_{\rm eff}^{\min}
 =
 \min_{0\le j\le p}n_{{\rm eff},j}.
 \label{eq:neff}
\end{equation}
For a joint block with alphabet size \(K_B\), using the attenuation factor \(\kappa_{\rm rr}\left(K_B,\eps\right)\) defined in \eqref{eq:block-rr-contraction}, the analogous coordinate-level index is
\[
 n_{{\rm eff},j}^{\rm blk}
 =
 n\,\E_B\left\{\ind\left(j\in B\right)\kappa_{\rm rr}\left(K_B,\eps\right)^2\right\},
 \qquad
 n_{\rm eff}^{{\rm blk},\min}
 =
 \min_{0\le j\le p}n_{{\rm eff},j}^{\rm blk}.
\]
These indices do not replace the covariance \(\Omega_\eps\), but they help diagnose slow convergence when \(\eps\) is small, the selected block is sparse, or the randomized-response alphabet is large.

Finally, the nonprivate online QR recursion is recovered along high-privacy-budget and high-resolution design paths.

\begin{corollary}[High-\(\eps\) recovery]
\label{cor:high-eps}
Consider a public design path
\(\mathsf{CQ}_X\left(q_\eps,s_\eps\right)\) such that \(s_\eps=p+1\) eventually,
\(q_\eps\to\infty\), and $
 \kappa_{\rm rr}\left(K_\eps,\eps\right)\to1,
 K_\eps=2q_\eps^p $
once \(s_\eps=p+1\).  Then, for each fixed \(\beta\) in a neighborhood of \(\bstar\),
$
 \gtilde_{\theta_\eps}\left(\beta\right)\to g_\beta\left(\mathcal O\right)$
in \(L^2\).

Under Condition~\ref{cond:population-qr}, the convergence is locally uniform over fixed compact neighborhoods of \(\bstar\).  Moreover, $
 \Omega_\eps\to\Omega_0
 =
 \Var\left\{g_{\bstar}\left(\mathcal O\right)\right\}.$

If the linear conditional quantile model is correctly specified, so that
\(\Pr\left(Y\le X^\top\bstar\mid X\right)=\tau\) almost surely, then $
 \Omega_0=\tau\left(1-\tau\right)\E\left(XX^\top\right).$
Consequently, the limiting covariance in Theorem~\ref{thm:an} approaches the nonprivate ASGD covariance along this path.
\end{corollary}

Thus the fixed finite-channel theory describes the private regime, while the corollary clarifies the connection to ordinary nonprivate online QR.  At small \(\eps\), randomized-response inversion inflates \(\Omega_\eps\); at large \(\eps\), with sufficiently refined all-coordinate quantization, the decoded input approaches the exact QR estimating-equation contribution.

\section{Numerical Experiments}
\label{sec:empirical-study}
This section evaluates the finite-sample accuracy and inference behavior of the locally private online estimator in controlled simulations.  The comparisons focus on the privacy path toward the nonprivate ASGD oracle, the gain from the support-aware mean-preserving channel \(\mathsf{CQ}_X\left(q,s\right)\), and the calibration of trajectory-based confidence summaries.

All methods use the same streaming data-generating model and projected Polyak-Ruppert recursion. The nonprivate ASGD estimator serves only as an oracle reference, using unperturbed estimating-equation inputs with no local quantization, coordinate selection, or randomized response; finite privacy budgets use \(\mathsf{CQ}_X\left(q,s\right)\).

\subsection{Simulation Setup}
\label{subsec:simulation-design}

For \(p\) non-intercept covariates, generate $
 X_i=\left(1,W_i^\top\right)^\top,
 W_{ij}\stackrel{\mathrm{ind}}{\sim}\mathrm{Unif}\left[-1,1\right],
 j=1,\ldots,p.$
The target coefficient is
\[
 \bstar
 =
 \frac{\left(1,-2,3,-4,\ldots,\left(-1\right)^p\left(p+1\right)\right)^\top}
 {\bigl\|\left(1,-2,3,-4,\ldots,\left(-1\right)^p\left(p+1\right)\right)^\top\bigr\|_2}.
\]
The alternating signs and increasing magnitudes make the slope effects heterogeneous across coordinates, while the normalization keeps the signal scale fixed across \(p\).
Conditional on \(X_i\), responses are generated by
\[
 Y_i
 =
 X_i^\top\bstar
 +
 0.5\left\{\zeta_i-\log\left(\tau/\left(1-\tau\right)\right)\right\},
 \qquad
 \zeta_i\sim\mathrm{Logistic}\left(0,1\right).
\]
Because the \(\tau\)-quantile of \(\zeta_i\) is \(\log\left(\tau/\left(1-\tau\right)\right)\), the conditional \(\tau\)-quantile of \(Y_i\) given \(X_i\) is \(X_i^\top\bstar\). The scale \(0.5\) fixes the noise level and is held constant across all simulation cells.

The public simulation grid is $
 \tau\in\left\{0.5,0.75\right\},
 p\in\left\{1,2,5,8\right\},
 \eps\in\left\{0.5,1,2,4,8,16,\infty\right\}.$
We use value \(\eps=\infty\) to denote the separate nonprivate ASGD oracle.

For each finite-\(\eps\) cell, the local channel is \(\mathsf{CQ}_X\left(q,s\right)\), with \(\left(q,s\right)\) selected before final evaluation from a prespecified coordinate-quantized candidate family using independent screening runs with the same synthetic model and step-size template. The selected choices are fixed for final evaluation and reported in Table~\ref{tab:channel-choices}; they are sparse at stringent privacy budgets and become richer through more selected coordinates or finer quantization as \(\eps\) increases.

\begin{table}[ht]
\centering
\setlength{\tabcolsep}{4pt}
\begin{tabular}{cccccccc}
\toprule
\(\tau\) & \(p\) & \(\eps=0.5\) & \(\eps=1\) & \(\eps=2\) & \(\eps=4\) & \(\eps=8\) & \(\eps=16\) \\
\midrule
0.50 & 1 & (4,1) & (2,1) & (4,1) & (4,2) & (16,2) & (8,2) \\
0.50 & 2 & (2,1) & (2,1) & (2,1) & (4,3) & (8,3) & (8,3) \\
0.50 & 5 & (2,1) & (2,1) & (4,1) & (8,1) & (3,5) & (4,6) \\
0.50 & 8 & (2,1) & (2,1) & (4,1) & (3,2) & (3,5) & (4,9) \\
0.75 & 1 & (4,1) & (2,2) & (2,2) & (4,2) & (8,2) & (16,2) \\
0.75 & 2 & (2,1) & (2,2) & (4,1) & (3,3) & (8,3) & (16,3) \\
0.75 & 5 & (2,1) & (4,1) & (4,1) & (3,2) & (3,6) & (8,6) \\
0.75 & 8 & (2,1) & (2,1) & (3,1) & (8,1) & (3,5) & (4,9) \\
\bottomrule
\end{tabular}
\caption{Private finite-\(\eps\) local-channel choices for the \(\mathsf{CQ}_X\left(q,s\right)\) simulations. Each entry reports \(\left(q,s\right)\), where \(q\) is the slope-grid size and \(s\) is the selected estimating-equation coordinate-block size.}
\label{tab:channel-choices}
\end{table}

\subsection{Online Recursion and Reported Summaries}
\label{subsec:simulation-targets}
The numerical implementation follows Algorithm~\ref{alg:private-online-qr}. For private finite-\(\eps\) cells, \(\gtilde_i\left(\beta_{i-1}\right)\) is the decoded private estimating-equation input produced by the coordinate-quantized channel. For the nonprivate oracle cells, the reference implementation replaces this input by the unperturbed estimating-equation contribution \(g_{\beta_{i-1}}\left(\mathcal O_i\right)\), with no local quantization, coordinate selection, or randomized response. All synthetic runs initialize at \(\beta_0=0\) and use the same coordinate-wise projection set \([-10,10]^{p+1}\), including the geometric baseline runs.

The simulations use the public stabilized power schedule \(\eta_i=r_{\mathcal M}/(i^{0.65}+300)\), with method-specific positive scale \(r_{\mathcal M}\) fixed before the stream begins. This finite-sample version of the Polyak-Ruppert schedule in Condition~\ref{cond:asgd-schedule} uses public scales recorded in \appComputational{} for replication.

Point-estimation summaries are recorded at the reporting set
\[
 \mathcal T=
 \left\{10^3,3{\times}10^3,10^4,3{\times}10^4,
 10^5,3{\times}10^5,10^6,3{\times}10^6,
 10^7,3{\times}10^7,10^8,3{\times}10^8\right\}.
\]
A simulation cell fixes \(\left(\tau,p,\eps\right)\) and, for \(\eps<\infty\), the public channel choice \(\left(q,s\right)\). The \(48\) private finite-\(\eps\) \(\mathsf{CQ}_X\left(q,s\right)\) cells and \(8\) nonprivate oracle cells use \(500\) independent Monte Carlo replications per cell, while the two geometric baselines use \(200\) replications over the same \(48\) private cells. Estimation performance is measured by \(\|\bar\beta_n-\bstar\|_2\), and uncertainty quantification by empirical coverage of nominal \(95\%\) coordinate intervals, dense-contrast intervals for \(a_{\rm den}^\top\bstar\) with \(a_{\rm den}=(p+1)^{-1/2}\mathbf 1_{p+1}\), and coefficient ellipsoids. We reserve \(\tau\) for the quantile level and use \(\alpha_{\rm ci}\) for inferential miscoverage, so all intervals and ellipsoids have nominal level \(1-\alpha_{\rm ci}=0.95\).

Because each user contributes only one privatized category, nominal stream length differs from usable information: coordinate selection lowers observation frequency for each block, and randomized-response inversion amplifies variation when \(\eps\) is small. Table~\ref{tab:effective-rootn-diagnostic} therefore reports representative block-level effective-information values and the observed large-\(n\) error decay of the averaged recursion, estimated by fitting \(\log\{\text{mean }\ell_2\text{ error}\}=a+b\log n\) over \(n\ge 10^6\). Across all \(48\) finite-\(\eps\) \(\mathsf{CQ}_X(q,s)\) cells, the fitted slopes have median \(-0.503\), IQR \(-0.509\) to \(-0.500\), and range \(-0.531\) to \(-0.491\), so stringent privacy mainly changes the effective-information constant while fixed-channel averaged trajectories retain root-\(n\)-type decay.

\begin{table}[H]
\centering

\setlength{\tabcolsep}{3.5pt}
\begin{tabular}{cccccccc}
\toprule
\(\left(\tau,p,\eps\right)\) & \(\left(q,s\right)\) & \(n_{\rm eff}^{{\rm blk},\min}\) & \(10^6\) & \(10^7\) & \(10^8\) & \(3{\times}10^8\) & slope \(b\) \\
\midrule
\(\left(0.50,2,2\right)\) & \(\left(2,1\right)\) & \(5.80{\times}10^7\) & \(9.26{\times}10^{-3}\) & \(2.81{\times}10^{-3}\) & \(8.98{\times}10^{-4}\) & \(5.25{\times}10^{-4}\) & \(-0.503\) \\
\(\left(0.75,8,0.5\right)\) & \(\left(2,1\right)\) & \(2.00{\times}10^6\) & \(2.46{\times}10^{-1}\) & \(6.70{\times}10^{-2}\) & \(2.07{\times}10^{-2}\) & \(1.17{\times}10^{-2}\) & \(-0.531\) \\
\(\left(0.50,8,16\right)\) & \(\left(4,9\right)\) & \(2.91{\times}10^8\) & \(5.55{\times}10^{-3}\) & \(1.71{\times}10^{-3}\) & \(5.47{\times}10^{-4}\) & \(3.15{\times}10^{-4}\) & \(-0.502\) \\
\(\left(0.75,8,16\right)\) & \(\left(4,9\right)\) & \(2.91{\times}10^8\) & \(6.67{\times}10^{-3}\) & \(2.01{\times}10^{-3}\) & \(6.31{\times}10^{-4}\) & \(3.69{\times}10^{-4}\) & \(-0.505\) \\
\bottomrule
\end{tabular}
\caption{Effective-information and large-\(n\) trajectory diagnostic. The \(n_{\rm eff}^{{\rm blk},\min}\) column is computed at \(n=3\times10^8\) using the reference contraction in the public step-size rule. The four middle columns report mean Euclidean estimation error at the displayed stream lengths, and \(b\) is the fitted log-log slope over \(n\ge 10^6\).}
\label{tab:effective-rootn-diagnostic}
\end{table}

\subsection{Estimation Accuracy}
\label{subsec:estimation-accuracy}
The estimation summaries move from the privacy-accuracy path to mechanism comparisons. We first vary \(\eps\) at fixed dimension, then check whether the large-\(\eps\) behavior remains close to the nonprivate reference as dimension grows, and finally compare against geometric baselines at the final horizon.

At \(p=2\), the privacy path separates the proposed channel from both geometric releases.  Direct Laplace is the closest baseline at large \(\eps\) but remains above the matched \(\mathsf{CQ}_X(q,s)\) errors.  For \(\tau=0.75\), face-exponential release often drives small-\(\eps\) paths to the projection boundary because it preserves support without preserving the mean; the errors near \(9.8\) are therefore boundary-saturation failures rather than ordinary privacy-noise plateaus.

\begin{table}[H]
\centering

\setlength{\tabcolsep}{3pt}
\begin{tabular}{ccccccc}
\toprule
 & \multicolumn{2}{c}{\(\mathsf{CQ}_X(q,s)\)} & \multicolumn{2}{c}{Direct Laplace} & \multicolumn{2}{c}{Face-exponential} \\
\cmidrule(lr){2-3}\cmidrule(lr){4-5}\cmidrule(lr){6-7}
\(\eps\) & \(\tau=0.50\) & \(\tau=0.75\) & \(\tau=0.50\) & \(\tau=0.75\) & \(\tau=0.50\) & \(\tau=0.75\) \\
\midrule
0.5 & \(1.63{\times}10^{-3}\) & \(3.08{\times}10^{-3}\) & \(3.57{\times}10^{-3}\) & \(5.35{\times}10^{-3}\) & \(7.72{\times}10^{-3}\) & \(9.85^{\times}\) \\
1 & \(8.30{\times}10^{-4}\) & \(1.59{\times}10^{-3}\) & \(1.99{\times}10^{-3}\) & \(2.63{\times}10^{-3}\) & \(3.61{\times}10^{-3}\) & \(9.85^{\times}\) \\
2 & \(5.25{\times}10^{-4}\) & \(7.26{\times}10^{-4}\) & \(9.50{\times}10^{-4}\) & \(1.22{\times}10^{-3}\) & \(1.99{\times}10^{-3}\) & \(9.84^{\times}\) \\
4 & \(2.57{\times}10^{-4}\) & \(2.83{\times}10^{-4}\) & \(4.84{\times}10^{-4}\) & \(6.32{\times}10^{-4}\) & \(9.37{\times}10^{-4}\) & \(9.81^{\times}\) \\
8 & \(1.50{\times}10^{-4}\) & \(1.82{\times}10^{-4}\) & \(2.88{\times}10^{-4}\) & \(3.41{\times}10^{-4}\) & \(4.91{\times}10^{-4}\) & \(2.05{\times}10^{-1}\) \\
16 & \(1.41{\times}10^{-4}\) & \(1.67{\times}10^{-4}\) & \(1.87{\times}10^{-4}\) & \(2.29{\times}10^{-4}\) & \(2.70{\times}10^{-4}\) & \(6.74{\times}10^{-3}\) \\
\midrule
\(\infty\) & \(1.39{\times}10^{-4}\) & \(1.67{\times}10^{-4}\) & -- & -- & -- & -- \\
\bottomrule
\end{tabular}
\caption{Mean Euclidean estimation error at \(n=3\times10^8\) with \(p=2\), varying the privacy budget and release mechanism. The \(\mathsf{CQ}_X(q,s)\) finite-\(\eps\) rows use the local-channel choices in Table~\ref{tab:channel-choices}. The \(\eps=\infty\) row is the common nonprivate ASGD oracle and is not a Laplace or face-exponential release. The \(\mathsf{CQ}_X(q,s)\) means use \(500\) replications per cell, while the two geometric baselines use \(200\) replications per cell. A superscript \(\times\) marks a face-exponential value whose magnitude is dominated by saturation at the projection boundary, rather than by an ordinary privacy-noise plateau.}
\label{tab:p2-privacy-estimation}
\end{table}

Table~\ref{tab:dimension-large-eps-estimation} fixes \(\eps=16\) and asks whether the high-budget private channel is already close to the nonprivate ASGD reference as \(p\) varies. In these displayed fixed-dimensional cells, the ratios are near one for \(p\le 5\) and about \(1.12\) at \(p=8\). Thus the later baseline gaps should not be read as a weak proposed reference: at this large budget, \(\mathsf{CQ}_X(q,s)\) is already tracking the ASGD oracle closely.

\begin{table}[t]
\centering
\begin{tabular}{ccccc}
\toprule
\(\tau\) & \(p\) & \(\mathsf{CQ}_X(q,s)\), \(\eps=16\) & nonprivate ASGD & CQ/ASGD \\
\midrule
0.50 & 1 & \(1.03{\times}10^{-4}\) & \(1.01{\times}10^{-4}\) & 1.03 \\
0.50 & 2 & \(1.41{\times}10^{-4}\) & \(1.39{\times}10^{-4}\) & 1.02 \\
0.50 & 5 & \(2.45{\times}10^{-4}\) & \(2.25{\times}10^{-4}\) & 1.09 \\
0.50 & 8 & \(3.15{\times}10^{-4}\) & \(2.82{\times}10^{-4}\) & 1.12 \\
\midrule
0.75 & 1 & \(1.16{\times}10^{-4}\) & \(1.16{\times}10^{-4}\) & 1.00 \\
0.75 & 2 & \(1.67{\times}10^{-4}\) & \(1.67{\times}10^{-4}\) & 1.00 \\
0.75 & 5 & \(2.60{\times}10^{-4}\) & \(2.53{\times}10^{-4}\) & 1.03 \\
0.75 & 8 & \(3.69{\times}10^{-4}\) & \(3.29{\times}10^{-4}\) & 1.12 \\
\bottomrule
\end{tabular}
\caption{High-budget recovery check at \(n=3\times10^8\) with \(\eps=16\), varying \(p\). Rows cover \(\tau\in\left\{0.50,0.75\right\}\) and \(p\in\left\{1,2,5,8\right\}\). The \(\mathsf{CQ}_X(q,s)\) rows use the local-channel choices in Table~\ref{tab:channel-choices}. Ratios compare the private finite-\(\eps\) estimator with the nonprivate ASGD oracle. Each mean is computed over \(500\) independent Monte Carlo replications.}
\label{tab:dimension-large-eps-estimation}
\end{table}

Across the \(56\) final-horizon \(\mathsf{CQ}_X(q,s)\) and nonprivate configurations, Monte Carlo error is small relative to the reported means: the median Monte Carlo standard error is \(7.2\times10^{-6}\), or \(1.7\%\) of the corresponding mean error, and the largest relative value is \(3.2\%\). The intermediate-horizon behavior is summarized by the effective-information/root-\(n\) diagnostic in Table~\ref{tab:effective-rootn-diagnostic}.

The comparison with geometric baselines should be read at the decoded online-update level, not as a claim that every decoded full vector lies in the original two-face support \(\mathcal G_\tau\). Conditional on the public block \(B\), \(\mathsf{CQ}_X\left(q,s\right)\) uses the projected support geometry for selected coordinates and then applies an affine randomized-response inverse; this preserves the selected-block mean, and public block sampling recovers the full estimating-equation mean after Horvitz-Thompson reconstruction. Thus, when \(s<p+1\), the tested advantage is support-aware finite-alphabet privatization with mean-preserving decoding, rather than literal full-dimensional support preservation.

Table~\ref{tab:geometry-baseline-estimation} keeps the same matched baseline comparison but reports where the ratios arise. For each \(\left(\tau,p\right)\), the displayed ratio summaries are computed over the six finite privacy budgets at \(n=3\times10^8\); each ratio is the baseline mean Euclidean error divided by the matched \(\mathsf{CQ}_X(q,s)\) mean Euclidean error. A ratio greater than one favors the coordinate-quantized decoded estimator.

\begin{table}[H]
\centering

\setlength{\tabcolsep}{3.5pt}
\begin{tabular}{cccccc}
\toprule
\(\tau\) & \(p\) & \multicolumn{2}{c}{Direct Laplace/\(\mathsf{CQ}_X\)} & \multicolumn{2}{c}{Face-exponential/\(\mathsf{CQ}_X\)} \\
\cmidrule(lr){3-4}\cmidrule(lr){5-6}
 & & median & range & median & range \\
\midrule
0.50 & 1 & 1.65 & 1.09--1.89 & 3.07 & 1.39--3.86 \\
0.50 & 2 & 1.90 & 1.32--2.39 & 3.71 & 1.91--4.74 \\
0.50 & 5 & 2.86 & 1.81--3.35 & 5.50 & 3.37--6.90 \\
0.50 & 8 & 3.67 & 2.57--4.14 & 7.45 & 5.07--8.76 \\
\midrule
0.75 & 1 & 1.52 & 1.10--2.17 & \(5.70{\times}10^3{}^{\times}\) & \(68.5\)--\(2.93{\times}10^4{}^{\times}\) \\
0.75 & 2 & 1.71 & 1.37--2.23 & \(4.69{\times}10^3{}^{\times}\) & \(40.2\)--\(3.46{\times}10^4{}^{\times}\) \\
0.75 & 5 & 3.04 & 2.73--3.54 & \(4.21{\times}10^3{}^{\times}\) & \(38.4\)--\(2.49{\times}10^4{}^{\times}\) \\
0.75 & 8 & 4.01 & 3.52--4.19 & \(2.75{\times}10^3{}^{\times}\) & \(26.8\)--\(1.52{\times}10^4{}^{\times}\) \\
\bottomrule
\end{tabular}
\caption{Matched estimation-only comparison with geometric baselines at \(n=3\times10^8\). The entries are baseline-to-\(\mathsf{CQ}_X(q,s)\) mean-error ratios, not raw errors: the Direct Laplace columns compare Direct Laplace against \(\mathsf{CQ}_X(q,s)\), and the Face-exponential columns compare Face-exponential against \(\mathsf{CQ}_X(q,s)\). For each \(\left(\tau,p\right)\), medians and ranges summarize these ratios over \(\eps\in\{0.5,1,2,4,8,16\}\). Baseline configurations use \(200\) independent replications, while the \(\mathsf{CQ}_X(q,s)\) comparison values use \(500\) independent replications. A superscript \(\times\) marks a face-exponential aggregate dominated by cells whose reported errors are governed by projection-boundary saturation.}
\label{tab:geometry-baseline-estimation}
\end{table}
Over all \(48\) matched finite-\(\eps\) cells, the median ratios are \(2.31\) for Direct Laplace and \(17.8\) for Face-exponential, with every ratio above one. Direct Laplace is the more stable comparator but its relative cost increases with dimension, whereas Face-exponential is only moderately worse for \(\tau=0.50\) but orders of magnitude worse for \(\tau=0.75\), consistent with mean-field mismatch rather than ordinary privacy-noise inflation.

\subsection{Uncertainty Quantification}
\label{subsec:uncertainty-quantification}

The coverage experiments ask whether uncertainty can be estimated from the same locally private online trajectory. The primary procedure is SN, calibrated by Theorem~\ref{thm:sn}; we also report DC and HiGrad, both post-processing procedures based on decoded online estimates that require no additional access to raw data.

The inferential targets are coordinate intervals, the dense contrast \(a_{\rm den}^\top\bstar\), and coefficient ellipsoids at \(n=3\times10^8\). DC uses \(R_{\rm DC}=p+5\) disjoint groups, with \(t_{R_{\rm DC}-1}\) critical values for scalar intervals and the corresponding Hotelling form for vector ellipsoids when the group covariance is full rank. HiGrad is reported for public branching vectors \(\mathbf b_{\rm HG}=\left[4,4\right], \left[4,8\right],\) and \(\left[4,4,4\right]\), with \(16\), \(32\), and \(64\) leaves; because leaf estimates share public prefixes, studentization uses the known leaf-correlation template, and ellipsoid summaries are reported only when the covariance-rank condition holds. For nominal \(95\%\) SN scalar procedures, the Brownian-bridge critical value is \(6.7134\); for SN coefficient ellipsoids with \(p+1=2,3,6,9\), the critical values are \(103.584,174.998,464.956,864.991\), computed as Monte Carlo \(0.95\) quantiles of the Brownian-bridge quadratic-form limits using \(200000\) paths, \(1000\) grid points, and a fixed seed.

Table~\ref{tab:p2-eps2-coverage} reports a moderate private configuration, \(p=2\) and \(\eps=2\), at the final horizon. The three HiGrad trees are shown separately because the public tree determines the finite-sample studentization.

\begin{table}[t]
\centering

\begin{tabular}{ccccc}
\toprule
\(\tau\) & Method & coordinate & dense & ellipsoid \\
\midrule
0.50 & SN & 0.934-0.946 & 0.940 & 0.934 \\
0.50 & DC & 0.948-0.966 & 0.950 & 0.954 \\
0.50 & HiGrad \(\left[4,4\right]\) & 0.934-0.950 & 0.956 & 0.952 \\
0.50 & HiGrad \(\left[4,8\right]\) & 0.946-0.952 & 0.954 & 0.952 \\
0.50 & HiGrad \(\left[4,4,4\right]\) & 0.932-0.940 & 0.936 & 0.934 \\
\midrule
0.75 & SN & 0.952-0.960 & 0.962 & 0.946 \\
0.75 & DC & 0.884-0.952 & 0.948 & 0.936 \\
0.75 & HiGrad \(\left[4,4\right]\) & 0.932-0.944 & 0.934 & 0.920 \\
0.75 & HiGrad \(\left[4,8\right]\) & 0.916-0.922 & 0.918 & 0.914 \\
0.75 & HiGrad \(\left[4,4,4\right]\) & 0.904-0.940 & 0.926 & 0.896 \\
\bottomrule
\end{tabular}
\caption{Final empirical \(95\%\) coverage at \(p=2\), \(\eps=2\), and \(n=3\times10^8\). SN denotes trajectory self-normalization, DC denotes divide-and-conquer disjoint-group studentization, and HiGrad denotes hierarchical group studentization. Coordinate columns report the min-max over the \(p+1=3\) coordinate intervals; dense denotes the \(a_{\rm den}=(p+1)^{-1/2}\mathbf 1_{p+1}\) contrast; ellipsoid denotes full-coefficient coverage. For HiGrad, \(\mathbf b_{\rm HG}\) denotes the public branching vector; $
 \mathbf b_{\rm HG}=\left[4,4\right], \left[4,8\right],$ and $\left[4,4,4\right],$ have \(R_{\rm HG}=16\), \(32\), and \(64\) leaves, respectively. Each displayed coverage estimate is computed from \(500\) independent Monte Carlo replications, giving Monte Carlo standard error at most \(0.023\).}
\label{tab:p2-eps2-coverage}
\end{table}

For \(\tau=0.50\), all reported methods are close to nominal coverage for coordinate intervals, the dense contrast, and coefficient ellipsoids. For \(\tau=0.75\), SN remains close to nominal, while DC and HiGrad show some undercoverage for particular coordinate and ellipsoid summaries; dense-contrast coverage is more stable than the least favorable coordinate and full-vector summaries. The upper-quantile, higher-dimensional, low-privacy-budget configuration \(\left(\tau,p,\eps\right)=\left(0.75,8,0.5\right)\) strengthens the same pattern: scalar dense-contrast coverage remains comparatively stable, whereas coordinate intervals and coefficient ellipsoids are more sensitive to finite effective information and covariance conditioning. The reported ellipsoid rows satisfy the required covariance-rank conditions, so the degradation reflects finite-sample conditioning rather than omitted rank failures.

The main empirical message is simple: once the effective-information index is large enough, the decoded \(\mathsf{CQ}_X\left(q,s\right)\) recursion tracks the nonprivate ASGD reference and outperforms the geometric baselines. The inference summaries reinforce the contrast-specific stability described above.

\FloatBarrier

\section{Real-Data: New York City Taxi Trips}
\label{sec:real-data-taxi}

We next evaluate the online recursion on January 2024 New York City yellow-taxi trip records released by the New York City Taxi and Limousine Commission \citep{nyctlc2024triprecord}. Each record contains pickup and dropoff timestamps, pickup and dropoff taxi-zone identifiers, trip distance, passenger count, and fare components, making the stream a natural illustration of the one-record, one-report local protocol: a trip record is treated as the sensitive contribution, records are timestamped, and the server processes them in chronological order without repeated interaction with the same record-level contributor. The illustration should not be read as a person-level guarantee for individuals who may appear in multiple trip records.

\subsection{Data Construction}

The response is
\[
Y=\log\left\{1+\hbox{trip duration in minutes}\right\},
\]
where duration is computed from the pickup and dropoff timestamps. The downloaded January file contains \(2{,}964{,}624\) raw trip records. We retain trips with pickup time in January 2024, duration between \(1\) and \(180\) minutes, positive trip distance no larger than \(100\) miles, positive fare amount, passenger count in \(\left\{1,\ldots,6\right\}\), and nonmissing pickup and dropoff locations. This leaves \(n=2{,}713{,}434\) trips. Table~\ref{tab:realdata-summary} summarizes the retained stream.

\begin{table}[H]
\centering
\begin{tabular}{lc}
\toprule
Quantity & Value \\
\midrule
Raw January 2024 yellow-taxi records & \(2{,}964{,}624\) \\
Retained records after cleaning & \(2{,}713{,}434\) \\
Training / test split sizes & \(2{,}062{,}859\) / \(650{,}575\) \\
Random split seed & 2026061903 \\
Trip duration Q1/median/Q3 (minutes) & 7.25 / 11.65 / 18.68 \\
Trip distance Q1/median/Q3 (miles) & 1.01 / 1.70 / 3.13 \\
Mean passenger count & 1.35 \\
Weekend pickups & 25.3\% \\
JFK/LGA pickup / dropoff & 8.2\% / 1.9\% \\
Manhattan pickups & 89.7\% \\
\bottomrule
\end{tabular}
\caption{Summary of the retained New York City yellow-taxi stream used in the real-data illustration. Q1 and Q3 denote the first and third empirical quartiles.}
\label{tab:realdata-summary}
\end{table}

The covariate vector has \(p=8\) non-intercept coordinates: clipped and rescaled \(\log\left(1+\hbox{trip distance}\right)\), sine and cosine transforms of pickup hour, a weekend indicator, passenger count scaled from \(1,\ldots,6\) to \(\left[-1,1\right]\), JFK/LGA airport pickup and dropoff indicators, and a Manhattan pickup indicator. The distance feature is clipped at \(20\) miles before scaling, all non-intercept coordinates lie in \(\left[-1,1\right]\), and the resulting target is a descriptive streaming linear QR projection for these clipped and publicly scaled features rather than a causal model for trip duration.

\subsection{Holdout Calibration and Stability}

Because the real-data population coefficient is unknown, we evaluate holdout calibration rather than coefficient coverage.  We sample \(650{,}575\) retained trips as a uniform holdout and process the remaining \(2{,}062{,}859\) trips once in pickup-time order.  For each \(\widehat\beta_\tau\), we report the hit rate
\[
 \widehat c_\tau
 =
 n_{\rm test}^{-1}\sum_{i\in{\rm test}}
 \ind\left(Y_i\le X_i^\top\widehat\beta_\tau\right),
\]
and the test check loss.  The nonprivate reference hit rates, \(0.445\) for \(\tau=0.50\) and \(0.672\) for \(\tau=0.75\), indicate misspecification of the clipped linear projection and one-pass reference; the real-data question is whether private estimates preserve this reference calibration and loss as \(\eps\) varies.

We use the same \(p=8\) channel choices from Table~\ref{tab:channel-choices}, initialize at zero, set \(\eta_i=0.08i^{-0.75}\), and project coordinatewise onto \(\left[-8,8\right]\). Table~\ref{tab:realdata-holdout} reports \(20\) independent repetitions of public block selection, stochastic quantization, and randomized response for each private cell, giving the holdout hit rate, \(10^3\) times the excess holdout check loss relative to the nonprivate training-stream reference, and the Euclidean distance between the private Polyak average and that reference. Specifically, if \(L_{{\rm test},\tau}(\beta)=n_{\rm test}^{-1}\sum_{i\in{\rm test}}\rho_\tau(Y_i-X_i^\top\beta)\), the reported loss contrast is \(10^3\Delta_{{\rm loss}}\), where \(\Delta_{{\rm loss}}=L_{{\rm test},\tau}(\widehat\beta_\tau)-L_{{\rm test},\tau}(\widehat\beta_\tau^{\rm np})\) and \(\widehat\beta_\tau^{\rm np}\) is the nonprivate one-pass ASGD estimate trained on the same stream.

\begin{table}[t]
\centering
\begin{tabular}{ccccccc}
\toprule
\(\tau\) & \(\eps\) & \(\left(q,s\right)\) & \(n_{\rm eff}^{{\rm blk},\min}\) & test hit rate & \(10^3\Delta_{{\rm loss}}\) & \(\ell_2\) distance \\
\midrule
0.50 & 2 & (4,1) & \(8.67{\times}10^{4}\) & 0.450 (0.008) & 7.40 (43.31) & 0.553 (0.133) \\
0.50 & 4 & (3,2) & \(3.40{\times}10^{5}\) & 0.448 (0.008) & 8.92 (17.32) & 0.255 (0.068) \\
0.50 & 8 & (3,5) & \(1.01{\times}10^{6}\) & 0.446 (0.002) & -1.18 (8.07) & 0.123 (0.042) \\
0.50 & 16 & (4,9) & \(2.00{\times}10^{6}\) & 0.445 (0.000) & 0.04 (0.92) & 0.013 (0.007) \\
0.50 & \(\infty\) & -- & \(2.06{\times}10^{6}\) & 0.445 & 0.00 & 0.000 \\
\midrule
0.75 & 2 & (3,1) & \(1.06{\times}10^{5}\) & 0.675 (0.014) & 23.23 (47.24) & 0.609 (0.253) \\
0.75 & 4 & (8,1) & \(1.74{\times}10^{5}\) & 0.674 (0.012) & 12.65 (48.34) & 0.475 (0.165) \\
0.75 & 8 & (3,5) & \(1.01{\times}10^{6}\) & 0.673 (0.004) & 3.30 (11.07) & 0.149 (0.034) \\
0.75 & 16 & (4,9) & \(2.00{\times}10^{6}\) & 0.672 (0.001) & -0.42 (1.21) & 0.021 (0.012) \\
0.75 & \(\infty\) & -- & \(2.06{\times}10^{6}\) & 0.672 & 0.00 & 0.000 \\
\bottomrule
\end{tabular}
\caption{Holdout calibration and stability for New York City yellow-taxi trips. The test set is a uniform random holdout of \(650{,}575\) retained trips, drawn without replacement using public seed \(2026061903\); the remaining \(2{,}062{,}859\) trips form the training stream. Finite-\(\eps\) entries are means with Monte Carlo standard deviations in parentheses over \(20\) repetitions of the local-channel randomness. The \(\eps=\infty\) rows are the nonprivate one-pass ASGD references trained on the same training stream.}
\label{tab:realdata-holdout}
\end{table}

The holdout results mirror the simulations without requiring a known coefficient.  At \(\eps=2\) and \(4\), private estimates show larger coefficient and excess-loss variation while keeping average hit rates near the nonprivate reference.  At \(\eps=8\), excess test losses are close to zero, and at \(\eps=16\) the higher-resolution all-coordinate channel nearly matches the nonprivate stream reference, with hit rates agreeing to three decimals and mean \(\ell_2\) distances \(0.013\) and \(0.021\) for \(\tau=0.50\) and \(0.75\).  Thus larger \(\eps\) mainly improves preservation of nonprivate holdout calibration and predictive check loss.

We also ran the two geometric baselines from Section~\ref{sec:problem-setup} on the same real-data split, using the same training stream, holdout set, step-size rule, projection set, privacy budgets, and \(20\) repetitions per private cell. Table~\ref{tab:realdata-baseline-comparison} summarizes these \(8\) finite-\(\eps\) cells; the same mean/support tradeoff appears in the holdout diagnostics, with the face-exponential shift especially visible for \(\tau=0.75\).

\begin{table}[H]
\centering

\setlength{\tabcolsep}{4pt}
\begin{tabular}{lrrrr}
\toprule
Method & med. \(|\Delta_{{\rm hit}}|\) & med. \(10^3\Delta_{{\rm loss}}\) & med. \(\ell_2\) & max \(\ell_2\) \\
\midrule
\(\mathsf{CQ}_X\left(q,s\right)\) & 0.002 & 5.35 & 0.202 & 0.61 \\
Direct Laplace & 0.003 & 20.09 & 0.654 & 2.61 \\
Face-exponential & 0.328 & 753.56 & 3.683 & 5.93 \\
\bottomrule
\end{tabular}
\caption{Real-data comparison with geometric baselines over the \(8\) finite-\(\eps\) cells. Each entry is a median of cell means, where each private cell mean uses \(20\) repetitions of the corresponding local randomization. Here \(\Delta_{{\rm hit}}\) is the test hit rate minus the nonprivate holdout hit rate at the same \(\tau\); \(\Delta_{{\rm loss}}\) is the excess holdout check loss defined above; and the \(\ell_2\) distance is relative to the nonprivate training-stream reference.}
\label{tab:realdata-baseline-comparison}
\end{table}

Using the displayed \(\mathsf{CQ}_X(q,s)\) row as the calibration scale, the Direct Laplace medians are about \(1.8\), \(3.8\), \(3.2\), and \(4.3\) times as large for hit-rate deviation, test loss, median \(\ell_2\) distance, and maximum \(\ell_2\) distance. The corresponding Face-exponential multipliers are about \(208\), \(141\), \(18.2\), and \(9.7\).

\section{Concluding Remarks}
\label{sec:discussion}

This paper develops a one-report local-privacy framework for online quantile regression.  The key step is to view the usual QR update as an unobserved estimating-equation contribution and to replace it by a decoded LDP input with the same conditional mean.  The \(\mathsf{CQ}_X\left(q,s\right)\) channel combines the two-face QR support, stochastic quantization, randomized response, affine decoding, and Horvitz-Thompson reconstruction.  For fixed finite designs, this yields consistent projected Polyak-Ruppert estimation, a central limit theorem, and Hessian-free scalar inference.  The simulations and taxi-data illustration show the corresponding privacy-accuracy tradeoff and the benefits of preserving both support structure and mean unbiasedness.

The theory is deliberately fixed-dimensional, with bounded covariates, nonsingular local QR Hessian, public finite-channel choices, and one LDP report per participant.  These assumptions yield exact \(\eps\)-LDP for each participant's one-record contribution and a clean stochastic-approximation limit, while requiring public scaling or clipping and predictable adaptive choices.  Scalar and prespecified low-dimensional inference is the most stable; full ellipsoids are more sensitive to conditioning, rank, and finite effective information.

Extensions such as high-dimensional QR, adaptive channel design, and locally private Hessian or density estimation require new mechanisms, additional reports, stronger modeling assumptions, or new one-report constructions.  The main conclusion is that one-report local privacy is compatible with online QR when the report and decoder preserve the estimating-equation mean, at the cost of variance inflation under tighter privacy, larger alphabets, sparse coordinate selection, or higher dimension.

\section{Acknowledgments}
The authors acknowledge computational resources provided by the Digital Research Alliance of Canada. The authors report no additional funding for this work and no competing interests relevant to this submission.

\FloatBarrier

\appendix

\section{Additional Numerical-Experiment Details}
\label{app:simulation-supplement}
\label{app:simulation-specification}

This appendix records implementation details for the simulation study in Section~\ref{sec:empirical-study}. The main text reports the primary privacy path, high-budget recovery check, geometry-baseline comparison, and a representative coverage table. The table below gives the public run configuration used for the additional numerical summaries in Appendix~\ref{app:additional-experimental-summaries}.

\subsection{Public Design and Implementation}

The public simulation grid is
\[
 \tau\in\{0.5,0.75\},\qquad
 p\in\{1,2,5,8\},\qquad
 \eps\in\{0.5,1,2,4,8,16,\infty\}.
\]
The finite-\(\eps\) \(\mathsf{CQ}_X(q,s)\) cells use the public channel choices reported in Table~\ref{tab:channel-choices} of the main text. These choices were selected before final evaluation from an independent \(10^7\)-user screening profile over the same synthetic model and step-size template. The nonprivate \(\eps=\infty\) rows use unperturbed QR estimating-equation inputs, all coordinates, no local randomization, and no finite \((q,s)\) channel.

All synthetic streams are generated online. The final \(\mathsf{CQ}_X(q,s)\) summaries cover \(48\) finite-\(\eps\) private cells plus \(8\) nonprivate reference cells and use \(500\) independent Monte Carlo replications per cell. The estimation-only direct Laplace and face-exponential baseline summaries use \(200\) replications for each of the \(96\) baseline cells formed by the \(48\) private finite-\(\eps\) cells and the two baseline mechanisms. No raw synthetic records are stored as part of the reported summary tables.

The point-estimation summaries report sample sizes
\[
 \mathcal T=
 \{10^3,3{\times}10^3,10^4,3{\times}10^4,
 10^5,3{\times}10^5,10^6,3{\times}10^6,
 10^7,3{\times}10^7,10^8,3{\times}10^8\}.
\]
For private finite-channel cells, the ASGD recursion uses the public stabilized power schedule
\[
 \eta_i
 =
 \frac{10\,\kappa_{\rm ref}(\eps)\sqrt s/d}{i^{0.65}+300},
 \qquad d=p+1,
\]
where \(s=|B|\), \(K_{\rm ref}=\max_{B:|B|=s}K_B\), and \(\kappa_{\rm ref}(\eps)=\kappa_{\rm rr}(K_{\rm ref},\eps)\). The nonprivate reference uses the same template with all coordinates and \(\kappa_{\rm ref}(\infty)=1\). All runs initialize at zero and project coordinatewise onto \([-10,10]^d\).

The inference summaries use trajectory self-normalization (SN), divide-and-conquer studentization (DC), and hierarchical group studentization (HiGrad). DC uses \(R_{\rm DC}=p+5\) disjoint groups. HiGrad uses the three public branching vectors in Table~\ref{tab:higrad-tree-budget}. For the nominal \(95\%\) SN procedures, the scalar Brownian-bridge critical value is \(6.7134\), and the coefficient-ellipsoid SN critical values for \(d=2,3,6,9\) are \(103.584\), \(174.998\), \(464.956\), and \(864.991\), respectively. These are Monte Carlo \(0.95\) quantiles of the corresponding Brownian-bridge quadratic-form limits, computed with \(200000\) Brownian paths and \(1000\) grid points.

\begin{table}[!htbp]
\centering
\small
{\setlength{\tabcolsep}{3pt}
\begin{tabular}{@{}lcc@{}}
\toprule
Tree \(\mathbf b_{\rm HG}\) & node counts \((N_\ell^{\rm HG})\) & weights \((w_\ell^{\rm HG})\) \\
\midrule
\([4,4]\) & \(1,4,16\) & \(0.0476,0.1905,0.7619\) \\
\([4,8]\) & \(1,4,32\) & \(0.0270,0.1081,0.8649\) \\
\([4,4,4]\) & \(1,4,16,64\) & \(0.0118,0.0471,0.1882,0.7529\) \\
\bottomrule
\end{tabular}
}
\caption{HiGrad public tree budget at \(n=3\times10^8\). Here \(N_\ell^{\rm HG}\) is the number of nodes at level \(\ell\), and \(w_\ell^{\rm HG}\) is the fraction of users assigned to that level. In each row, the displayed weights sum to one up to rounding.}
\label{tab:higrad-tree-budget}
\end{table}

\FloatBarrier

\section{Additional Numerical Summaries}
\label{app:additional-experimental-summaries}

This appendix reports supplementary point-estimation and \(n=3\times10^8\) coverage summaries for the decoded estimating-equation experiments. Table~\ref{tab:sample-size-estimation} expands the sample-size trajectories behind the root-\(n\) diagnostic in the main text. Tables~\ref{tab:geometry-baseline-ratio-epsilon}, \ref{tab:geometry-baseline-ratio-p}, and \ref{tab:geometry-baseline-ratio-distribution} give alternative aggregations of the estimation-only direct Laplace and face-exponential comparisons. Tables~\ref{tab:hard-privacy-sensitivity}--\ref{tab:higrad-tree-coverage} summarize inference behavior in difficult cells and across the final-horizon grid. A single \(500\)-replication coverage estimate has Monte Carlo standard error at most \(0.023\).

\begin{table}[!htbp]
\centering

\begin{tabular}{rrrrr}
\toprule
\(n\) & \((0.5,2,2)\) & \((0.75,8,0.5)\) & \((0.75,8,16)\) & \((0.75,8,\infty)\) \\
\midrule
\(10^3\) & \(6.07{\times}10^{-1}\) & \(9.90{\times}10^{-1}\) & \(7.06{\times}10^{-1}\) & \(7.02{\times}10^{-1}\) \\
\(3{\times}10^3\) & \(3.20{\times}10^{-1}\) & \(9.86{\times}10^{-1}\) & \(4.26{\times}10^{-1}\) & \(4.20{\times}10^{-1}\) \\
\(10^4\) & \(1.25{\times}10^{-1}\) & \(9.53{\times}10^{-1}\) & \(1.63{\times}10^{-1}\) & \(1.59{\times}10^{-1}\) \\
\(3{\times}10^4\) & \(5.89{\times}10^{-2}\) & \(8.72{\times}10^{-1}\) & \(6.34{\times}10^{-2}\) & \(6.06{\times}10^{-2}\) \\
\(10^5\) & \(3.04{\times}10^{-2}\) & \(6.87{\times}10^{-1}\) & \(2.56{\times}10^{-2}\) & \(2.35{\times}10^{-2}\) \\
\(3{\times}10^5\) & \(1.71{\times}10^{-2}\) & \(4.62{\times}10^{-1}\) & \(1.29{\times}10^{-2}\) & \(1.15{\times}10^{-2}\) \\
\(10^6\) & \(9.26{\times}10^{-3}\) & \(2.46{\times}10^{-1}\) & \(6.67{\times}10^{-3}\) & \(5.96{\times}10^{-3}\) \\
\(3{\times}10^6\) & \(5.28{\times}10^{-3}\) & \(1.33{\times}10^{-1}\) & \(3.70{\times}10^{-3}\) & \(3.29{\times}10^{-3}\) \\
\(10^7\) & \(2.81{\times}10^{-3}\) & \(6.70{\times}10^{-2}\) & \(2.01{\times}10^{-3}\) & \(1.78{\times}10^{-3}\) \\
\(3{\times}10^7\) & \(1.62{\times}10^{-3}\) & \(3.86{\times}10^{-2}\) & \(1.17{\times}10^{-3}\) & \(1.04{\times}10^{-3}\) \\
\(10^8\) & \(8.98{\times}10^{-4}\) & \(2.07{\times}10^{-2}\) & \(6.31{\times}10^{-4}\) & \(5.63{\times}10^{-4}\) \\
\(3{\times}10^8\) & \(5.25{\times}10^{-4}\) & \(1.17{\times}10^{-2}\) & \(3.69{\times}10^{-4}\) & \(3.29{\times}10^{-4}\) \\
\bottomrule
\end{tabular}
\caption{Mean Euclidean estimation error at reported sample sizes \(n\in\{1,3\}\times10^{r_n}\), \(r_n=3,\ldots,8\). The columns fix \((\tau,p,\eps)=(0.5,2,2)\), \((0.75,8,0.5)\), \((0.75,8,16)\), and the nonprivate ASGD reference \((0.75,8,\infty)\). This table expands the trajectory information summarized in the main-text effective-information diagnostic. Each mean is computed over \(500\) independent Monte Carlo replications.}
\label{tab:sample-size-estimation}
\end{table}

\begin{table}[!htbp]
\centering

\begin{tabular}{rcc}
\toprule
\(\eps\) & Direct Laplace & Face-exponential \\
\midrule
0.5 & 2.58 & \(4.43{\times}10^2{}^{\times}\) \\
1 & 2.56 & \(8.20{\times}10^2{}^{\times}\) \\
2 & 2.48 & \(1.94{\times}10^3{}^{\times}\) \\
4 & 2.36 & \(3.49{\times}10^3{}^{\times}\) \\
8 & 2.28 & \(4.97{\times}10^2{}^{\times}\) \\
16 & 1.59 & 15.9 \\
\bottomrule
\end{tabular}
\caption{Median final-horizon estimation-error ratio by privacy budget for the geometric baselines. Each entry is the median, over \(\tau\in\{0.5,0.75\}\) and \(p\in\{1,2,5,8\}\), of the baseline mean Euclidean error divided by the corresponding \(\mathsf{CQ}_X(q,s)\) mean Euclidean error at \(n=3\times10^8\). A superscript \(\times\) marks a face-exponential aggregate dominated by cells whose reported errors are governed by projection-boundary saturation.}
\label{tab:geometry-baseline-ratio-epsilon}
\end{table}

\begin{table}[!htbp]
\centering
\begin{tabular}{rcc}
\toprule
\(p\) & Direct Laplace & Face-exponential \\
\midrule
1 & 1.54 & \(36.2^{\times}\) \\
2 & 1.84 & \(22.5^{\times}\) \\
5 & 3.02 & \(22.6^{\times}\) \\
8 & 3.81 & \(17.8^{\times}\) \\
\bottomrule
\end{tabular}
\caption{Median final-horizon estimation-error ratio by dimension for the geometric baselines. Each entry is the median, over \(\tau\in\{0.5,0.75\}\) and \(\eps\in\{0.5,1,2,4,8,16\}\), of the baseline mean Euclidean error divided by the corresponding \(\mathsf{CQ}_X(q,s)\) mean Euclidean error at \(n=3\times10^8\). A superscript \(\times\) marks a face-exponential aggregate dominated by cells whose reported errors are governed by projection-boundary saturation.}
\label{tab:geometry-baseline-ratio-p}
\end{table}

\begin{table}[!htbp]
\centering

\begin{tabular}{lrrrrl}
\toprule
Baseline & \(Q_{0.25}\) & median & \(Q_{0.75}\) & median MCSE & largest-ratio cell \\
\midrule
Direct Laplace & 1.76 & 2.31 & 3.37 & 0.078 & \((0.75,8,4)\): 4.19 \\
Face-exponential & 4.65 & \(17.8^{\times}\) & \(4.01{\times}10^3{}^{\times}\) & 0.261 & \((0.75,2,4)^{\times}\): \(3.46{\times}10^4\) \\
\bottomrule
\end{tabular}
\caption{Distribution of final-horizon estimation-error ratios for the geometric baselines over the \(48\) private finite-\(\eps\) cells. Ratios compare each baseline mean Euclidean error with the \(\mathsf{CQ}_X(q,s)\) mean in the same \((\tau,p,\eps)\) cell. The MCSE column is the median delta-method Monte Carlo standard error of the ratio, using \(200\) baseline replications and \(500\) \(\mathsf{CQ}_X(q,s)\) replications. A superscript \(\times\) marks a face-exponential aggregate or cell whose reported error is governed by projection-boundary saturation.}
\label{tab:geometry-baseline-ratio-distribution}
\end{table}

\begin{table}[!htbp]
\centering

\begin{tabular}{rccccccccc}
\toprule
\(\eps\) & \multicolumn{3}{c}{SN} & \multicolumn{3}{c}{DC} & \multicolumn{3}{c}{HiGrad \([4,4]\)} \\
\cmidrule(lr){2-4}\cmidrule(lr){5-7}\cmidrule(lr){8-10}
 & coord. 0 & dense & ell. & coord. 0 & dense & ell. & coord. 0 & dense & ell. \\
\midrule
0.5 & 0.900 & 0.934 & 0.586 & 0.006 & 0.942 & 0.684 & 0.648 & 0.868 & 0.746 \\
1 & 0.916 & 0.938 & 0.714 & 0.020 & 0.944 & 0.724 & 0.668 & 0.904 & 0.870 \\
2 & 0.946 & 0.954 & 0.840 & 0.492 & 0.950 & 0.904 & 0.878 & 0.938 & 0.920 \\
4 & 0.928 & 0.946 & 0.866 & 0.854 & 0.944 & 0.944 & 0.916 & 0.938 & 0.934 \\
8 & 0.946 & 0.960 & 0.916 & 0.744 & 0.946 & 0.928 & 0.922 & 0.956 & 0.944 \\
16 & 0.956 & 0.954 & 0.932 & 0.848 & 0.952 & 0.946 & 0.936 & 0.924 & 0.938 \\
\midrule
\(\infty\) & 0.956 & 0.950 & 0.904 & 0.886 & 0.944 & 0.942 & 0.930 & 0.944 & 0.938 \\
\bottomrule
\end{tabular}
\caption{Coverage sensitivity at \(n=3\times10^8\) in the upper-quantile larger-\(p\) design \((\tau,p)=(0.75,8)\). Finite-\(\eps\) rows vary \(\eps\); the \(\infty\) row is the nonprivate ASGD reference. The HiGrad columns use only the public tree \(\mathbf b_{\rm HG}=[4,4]\); Table~\ref{tab:higrad-fixed-cell-tree-coverage} isolates tree sensitivity. Coordinate entries refer to coordinate \(0\), dense denotes the \(a_{\rm den}=d^{-1/2}\mathbf 1_d\) contrast, and ellipsoid denotes full-coefficient coverage. Each displayed coverage estimate is based on \(500\) independent Monte Carlo replications.}
\label{tab:hard-privacy-sensitivity}
\end{table}

\begin{table}[!htbp]
\centering

\begin{tabular}{llrrr}
\toprule
\((\tau,p,\eps)\) & Estimand & \(\mathbf b_{\rm HG}=[4,4]\) & \(\mathbf b_{\rm HG}=[4,8]\) & \(\mathbf b_{\rm HG}=[4,4,4]\) \\
\midrule
\((0.50,2,2)\) & coord. 0 & 0.950 & 0.952 & 0.934 \\
\((0.50,2,2)\) & dense & 0.956 & 0.954 & 0.936 \\
\((0.50,2,2)\) & ellipsoid & 0.952 & 0.952 & 0.934 \\
\midrule
\((0.75,2,2)\) & coord. 0 & 0.932 & 0.916 & 0.904 \\
\((0.75,2,2)\) & dense & 0.934 & 0.918 & 0.926 \\
\((0.75,2,2)\) & ellipsoid & 0.920 & 0.914 & 0.896 \\
\midrule
\((0.75,8,0.5)\) & coord. 0 & 0.648 & 0.498 & 0.176 \\
\((0.75,8,0.5)\) & dense & 0.868 & 0.734 & 0.748 \\
\((0.75,8,0.5)\) & ellipsoid & 0.746 & 0.316 & 0.136 \\
\midrule
\((0.75,8,16)\) & coord. 0 & 0.936 & 0.932 & 0.914 \\
\((0.75,8,16)\) & dense & 0.924 & 0.940 & 0.938 \\
\((0.75,8,16)\) & ellipsoid & 0.938 & 0.918 & 0.920 \\
\bottomrule
\end{tabular}
\caption{HiGrad empirical \(95\%\) coverage at \(n=3\times10^8\) and fixed design points, varying only the public tree. Coordinate rows report coordinate \(0\), dense denotes the \(a_{\rm den}=d^{-1/2}\mathbf 1_d\) contrast, and ellipsoid denotes full-coefficient coverage. For HiGrad, \(\mathbf b_{\rm HG}\) denotes the public branching vector; \(\mathbf b_{\rm HG}=[4,4]\), \(\mathbf b_{\rm HG}=[4,8]\), and \(\mathbf b_{\rm HG}=[4,4,4]\) have \(R_{\rm HG}=16\), \(32\), and \(64\) leaves, respectively. Each displayed coverage estimate is based on \(500\) independent Monte Carlo replications.}
\label{tab:higrad-fixed-cell-tree-coverage}
\end{table}

\begin{table}[!htbp]
\centering

\begin{tabular}{llcccccc}
\toprule
\((\tau,p,\eps)\) & Method
& \multicolumn{2}{c}{coord. 0}
& \multicolumn{2}{c}{dense}
& \multicolumn{2}{c}{ellipsoid} \\
\cmidrule(lr){3-4}\cmidrule(lr){5-6}\cmidrule(lr){7-8}
& & cov. & mean length & cov. & mean length & cov. & mean volume \\
\midrule
\((0.50,2,2)\) & SN & 0.946 & \(6.48{\times}10^{-4}\) & 0.940 & \(1.65{\times}10^{-3}\) & 0.934 & \(5.74{\times}10^{-9}\) \\
 & DC & 0.948 & \(6.09{\times}10^{-4}\) & 0.950 & \(1.59{\times}10^{-3}\) & 0.954 & \(8.98{\times}10^{-9}\) \\
 & HiGrad \([4,4]\) & 0.950 & \(5.59{\times}10^{-4}\) & 0.956 & \(1.39{\times}10^{-3}\) & 0.952 & \(3.00{\times}10^{-9}\) \\
 & HiGrad \([4,8]\) & 0.952 & \(5.37{\times}10^{-4}\) & 0.954 & \(1.34{\times}10^{-3}\) & 0.952 & \(2.30{\times}10^{-9}\) \\
 & HiGrad \([4,4,4]\) & 0.934 & \(5.21{\times}10^{-4}\) & 0.936 & \(1.30{\times}10^{-3}\) & 0.934 & \(1.99{\times}10^{-9}\) \\
\midrule
\((0.75,8,0.5)\) & SN & 0.900 & \(5.28{\times}10^{-3}\) & 0.934 & \(1.91{\times}10^{-2}\) & 0.586 & \(1.03{\times}10^{-15}\) \\
 & DC & 0.006 & \(4.14{\times}10^{-3}\) & 0.942 & \(1.78{\times}10^{-2}\) & 0.684 & \(3.88{\times}10^{-13}\) \\
 & HiGrad \([4,4]\) & 0.648 & \(3.78{\times}10^{-3}\) & 0.868 & \(1.50{\times}10^{-2}\) & 0.746 & \(4.22{\times}10^{-15}\) \\
 & HiGrad \([4,8]\) & 0.498 & \(3.63{\times}10^{-3}\) & 0.734 & \(1.40{\times}10^{-2}\) & 0.316 & \(1.32{\times}10^{-16}\) \\
 & HiGrad \([4,4,4]\) & 0.176 & \(3.51{\times}10^{-3}\) & 0.748 & \(1.29{\times}10^{-2}\) & 0.136 & \(3.13{\times}10^{-17}\) \\
\bottomrule
\end{tabular}
\caption{Final-horizon interval and ellipsoid size for two representative cells at \(n=3\times10^8\). Coordinate and dense columns report empirical coverage and mean interval length. Ellipsoid columns report empirical coverage and mean ellipsoid volume. The first cell is the moderate \(p=2,\eps=2\) design used in main-text Table~\ref{tab:p2-eps2-coverage}; the second is the upper-quantile, larger-\(p\), small-\(\eps\) design discussed in Section~\ref{sec:empirical-study}. Each entry is computed from \(500\) independent Monte Carlo replications.}
\label{tab:fixed-cell-interval-size}
\end{table}

\begin{table}[!htbp]
\centering

{\setlength{\tabcolsep}{3pt}
\begin{tabular}{@{}rlccc@{}}
\toprule
\(\eps\) & Method & coordinate & dense & ellipsoid \\
\midrule
0.5 & SN & \(0.933\,[0.900,0.966]\) & \(0.934\,[0.912,0.960]\) & \(0.851\,[0.586,0.956]\) \\
 & DC & \(0.904\,[0.006,0.974]\) & \(0.950\,[0.938,0.964]\) & \(0.895\,[0.684,0.962]\) \\
\midrule
1 & SN & \(0.940\,[0.916,0.970]\) & \(0.950\,[0.938,0.968]\) & \(0.894\,[0.714,0.958]\) \\
 & DC & \(0.913\,[0.020,0.974]\) & \(0.948\,[0.942,0.956]\) & \(0.911\,[0.724,0.950]\) \\
\midrule
2 & SN & \(0.943\,[0.916,0.960]\) & \(0.946\,[0.928,0.962]\) & \(0.918\,[0.840,0.954]\) \\
 & DC & \(0.931\,[0.492,0.966]\) & \(0.949\,[0.926,0.972]\) & \(0.936\,[0.904,0.954]\) \\
\midrule
4 & SN & \(0.948\,[0.928,0.974]\) & \(0.952\,[0.946,0.958]\) & \(0.935\,[0.866,0.960]\) \\
 & DC & \(0.944\,[0.854,0.966]\) & \(0.949\,[0.930,0.958]\) & \(0.948\,[0.934,0.960]\) \\
\midrule
8 & SN & \(0.948\,[0.920,0.970]\) & \(0.953\,[0.934,0.962]\) & \(0.941\,[0.916,0.968]\) \\
 & DC & \(0.942\,[0.744,0.964]\) & \(0.946\,[0.934,0.956]\) & \(0.944\,[0.928,0.972]\) \\
\midrule
16 & SN & \(0.948\,[0.918,0.974]\) & \(0.951\,[0.942,0.962]\) & \(0.946\,[0.928,0.964]\) \\
 & DC & \(0.941\,[0.848,0.964]\) & \(0.945\,[0.932,0.952]\) & \(0.946\,[0.936,0.958]\) \\
\midrule
\(\infty\) & SN & \(0.948\,[0.916,0.978]\) & \(0.950\,[0.938,0.964]\) & \(0.940\,[0.904,0.956]\) \\
 & DC & \(0.942\,[0.886,0.966]\) & \(0.945\,[0.930,0.962]\) & \(0.944\,[0.930,0.954]\) \\
\bottomrule
\end{tabular}
}
\caption{Final \(95\%\) coverage at \(n=3\times10^8\) for SN and DC by privacy budget and estimand. Entries are mean empirical coverage with \([\min,\max]\) over the cells at the displayed \(\eps\), method, and estimand. For each \(\eps\)-method row, the coordinate entry summarizes \(40\) coordinate-specific coverages over \(\tau\in\{0.5,0.75\}\), \(p\in\{1,2,5,8\}\), and coordinates; the dense and ellipsoid entries summarize \(8\) cell-level coverages. Dense denotes the \(a_{\rm den}=d^{-1/2}\mathbf 1_d\) contrast, and ellipsoid denotes full-coefficient coverage. The \(\eps=\infty\) rows use the nonprivate ASGD reference. Each underlying coverage estimate is computed from \(500\) independent replications. HiGrad summaries are reported separately by tree in Table~\ref{tab:higrad-tree-coverage}.}
\label{tab:method-estimand-coverage}
\end{table}

\begin{table}[!htbp]
\centering

{\setlength{\tabcolsep}{3pt}
\begin{tabular}{@{}rllccc@{}}
\toprule
\(\eps\) & Tree & \(R_{\rm HG}\) & coordinate & dense & ellipsoid \\
\midrule
0.5 & \([4,4]\) & 16 & \(0.902\,[0.648,0.962]\) & \(0.922\,[0.868,0.948]\) & \(0.895\,[0.746,0.948]\) \\
 & \([4,8]\) & 32 & \(0.848\,[0.498,0.960]\) & \(0.875\,[0.734,0.922]\) & \(0.754\,[0.316,0.920]\) \\
 & \([4,4,4]\) & 64 & \(0.829\,[0.176,0.962]\) & \(0.876\,[0.748,0.932]\) & \(0.704\,[0.136,0.944]\) \\
\midrule
1 & \([4,4]\) & 16 & \(0.924\,[0.668,0.958]\) & \(0.937\,[0.904,0.954]\) & \(0.919\,[0.866,0.964]\) \\
 & \([4,8]\) & 32 & \(0.887\,[0.478,0.950]\) & \(0.910\,[0.874,0.940]\) & \(0.840\,[0.578,0.942]\) \\
 & \([4,4,4]\) & 64 & \(0.876\,[0.238,0.962]\) & \(0.910\,[0.842,0.944]\) & \(0.797\,[0.342,0.956]\) \\
\midrule
2 & \([4,4]\) & 16 & \(0.937\,[0.878,0.958]\) & \(0.941\,[0.908,0.966]\) & \(0.934\,[0.912,0.954]\) \\
 & \([4,8]\) & 32 & \(0.920\,[0.808,0.958]\) & \(0.926\,[0.890,0.954]\) & \(0.907\,[0.804,0.952]\) \\
 & \([4,4,4]\) & 64 & \(0.914\,[0.748,0.964]\) & \(0.924\,[0.906,0.940]\) & \(0.881\,[0.720,0.954]\) \\
\midrule
4 & \([4,4]\) & 16 & \(0.944\,[0.888,0.970]\) & \(0.951\,[0.938,0.966]\) & \(0.947\,[0.932,0.976]\) \\
 & \([4,8]\) & 32 & \(0.931\,[0.876,0.966]\) & \(0.939\,[0.900,0.976]\) & \(0.921\,[0.848,0.948]\) \\
 & \([4,4,4]\) & 64 & \(0.930\,[0.882,0.958]\) & \(0.931\,[0.908,0.946]\) & \(0.905\,[0.818,0.934]\) \\
\midrule
8 & \([4,4]\) & 16 & \(0.942\,[0.916,0.960]\) & \(0.948\,[0.936,0.958]\) & \(0.940\,[0.930,0.950]\) \\
 & \([4,8]\) & 32 & \(0.941\,[0.916,0.968]\) & \(0.944\,[0.898,0.978]\) & \(0.935\,[0.912,0.968]\) \\
 & \([4,4,4]\) & 64 & \(0.938\,[0.882,0.960]\) & \(0.942\,[0.926,0.954]\) & \(0.927\,[0.884,0.950]\) \\
\midrule
16 & \([4,4]\) & 16 & \(0.948\,[0.926,0.970]\) & \(0.948\,[0.924,0.964]\) & \(0.947\,[0.938,0.960]\) \\
 & \([4,8]\) & 32 & \(0.946\,[0.926,0.964]\) & \(0.946\,[0.916,0.966]\) & \(0.939\,[0.918,0.954]\) \\
 & \([4,4,4]\) & 64 & \(0.945\,[0.914,0.968]\) & \(0.942\,[0.934,0.960]\) & \(0.942\,[0.920,0.962]\) \\
\midrule
\(\infty\) & \([4,4]\) & 16 & \(0.945\,[0.922,0.974]\) & \(0.943\,[0.928,0.960]\) & \(0.941\,[0.922,0.952]\) \\
 & \([4,8]\) & 32 & \(0.944\,[0.920,0.968]\) & \(0.942\,[0.918,0.952]\) & \(0.942\,[0.912,0.956]\) \\
 & \([4,4,4]\) & 64 & \(0.939\,[0.900,0.960]\) & \(0.942\,[0.930,0.950]\) & \(0.929\,[0.906,0.950]\) \\
\bottomrule
\end{tabular}
}
\caption{HiGrad final \(95\%\) coverage by privacy budget and tree at \(n=3\times10^8\). Entries are mean empirical coverage with \([\min,\max]\) over the cells at the displayed \(\eps\), tree, and estimand. For each \(\eps\)-tree row, the coordinate entry summarizes \(40\) coordinate-specific coverages over \(\tau\in\{0.5,0.75\}\), \(p\in\{1,2,5,8\}\), and coordinates; the dense and ellipsoid entries summarize \(8\) cell-level coverages. Dense denotes the \(a_{\rm den}=d^{-1/2}\mathbf 1_d\) contrast, and ellipsoid denotes full-coefficient coverage. The \(\eps=\infty\) rows use the nonprivate ASGD reference. Each underlying coverage estimate is computed from \(500\) independent replications.}
\label{tab:higrad-tree-coverage}
\end{table}

\FloatBarrier

\bibliographystyle{abbrvnat}
\bibliography{bibliography/references}
\end{document}